\documentclass[letterpaper]{article} 
\usepackage{aaai23}  
\usepackage{times}  
\usepackage{helvet}  
\usepackage{courier}  
\usepackage[hyphens]{url}  
\usepackage{graphicx} 
\usepackage{natbib}  
\usepackage{caption} 
\usepackage{amssymb}
\usepackage{amsmath}
\usepackage{booktabs}
\usepackage{comment}
\usepackage{multirow}
\usepackage{subcaption}
\usepackage[ruled]{algorithm2e}
\usepackage{cleveref}
\crefformat{equation}{equation~#2#1#3}
\title{Behavior Estimation from Multi-Source Data for Offline Reinforcement Learning}
\author {
    Guoxi Zhang\textsuperscript{\rm 1} and
    Hisashi Kashima\textsuperscript{\rm 1,2}
}
\affiliations {
    \textsuperscript{\rm 1} Graduate School of Inforamtics, Kyoto University\\
    \textsuperscript{\rm 2} RIKEN Guardian Robot Project\\
    guoxi@ml.ist.i.kyoto-u.ac.jp, kashima@i.kyoto-u.ac.jp
}

\begin{document}

\maketitle

\begin{abstract}
Offline reinforcement learning (RL) have received rising interest due to its appealing data efficiency. The present study addresses behavior estimation, a task that aims at estimating the data-generating policy. In particular, this work considers a scenario where data are collected from multiple sources. Neglecting data heterogeneity, existing approaches cannot provide good estimates and impede policy learning. To overcome this drawback, the present study proposes a latent variable model and a model-learning algorithm to infer a set of policies from data, which allows an agent to use as behavior policy the policy that best describes a particular trajectory. To illustrate the benefit of such a fine-grained characterization for multi-source data, this work showcases how the proposed model can be incorporated into an existing offline RL algorithm. Lastly, with extensive empirical evaluation this work confirms the risks of neglecting data heterogeneity and the efficacy of the proposed model.
\end{abstract}

\section{Introduction}
\label{section:introduction}

In offline reinforcement learning (RL)~\citep{Lange2012}, agents learn policies using existing data without costly online interaction. Due to such data efficiency, offline RL becomes an intriguing paradigm for applications such as recommender systems~\citep{10.1145/3289600.3290999} and robot manipulation~\citep{pmlr-v100-dasari20a}.

\emph{Behavior estimation} refers to the task of computing an estimate for data-generating policy. Such an estimate is an essential component of many offline RL algorithms. For example, the discounted COP-TD algorithm~\citep{Gelada_Bellemare_2019} and the OPPOSD algorithm~\citep{pmlr-v115-liu20a} requires such estimate for importance sampling, and the BRAC-v algorithm~\citep{wu2019behavior} utilizes it in behavior regularization. Hence, behavior estimation is a premise for these algorithms to work in practice.

\begin{figure}[t!]
     \centering
     \begin{subfigure}[t]{0.48\columnwidth}
         \centering
         \includegraphics[width=\textwidth]{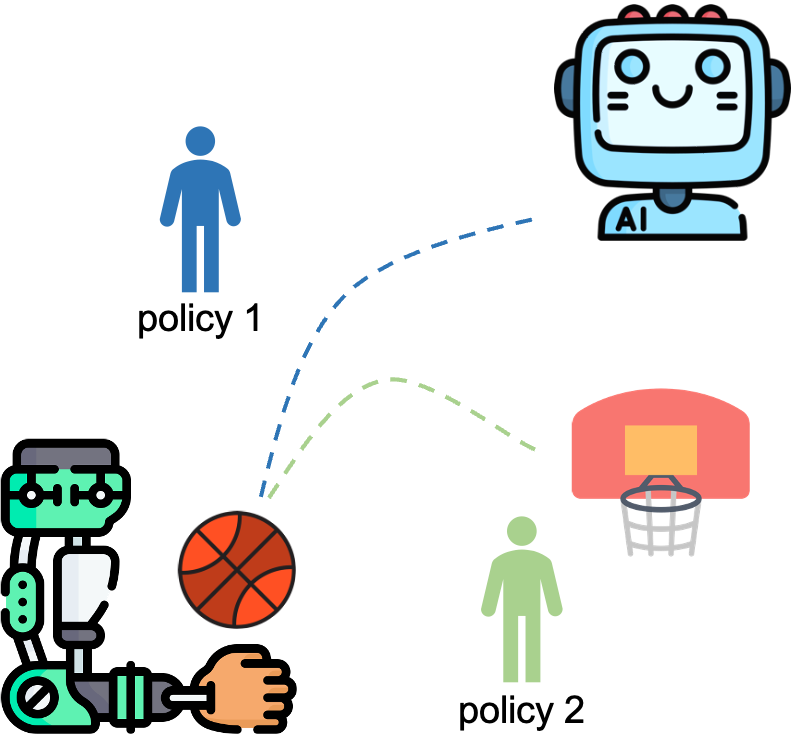}
     \end{subfigure}
     \begin{subfigure}[t]{0.48\columnwidth}
         \centering
         \includegraphics[width=\textwidth]{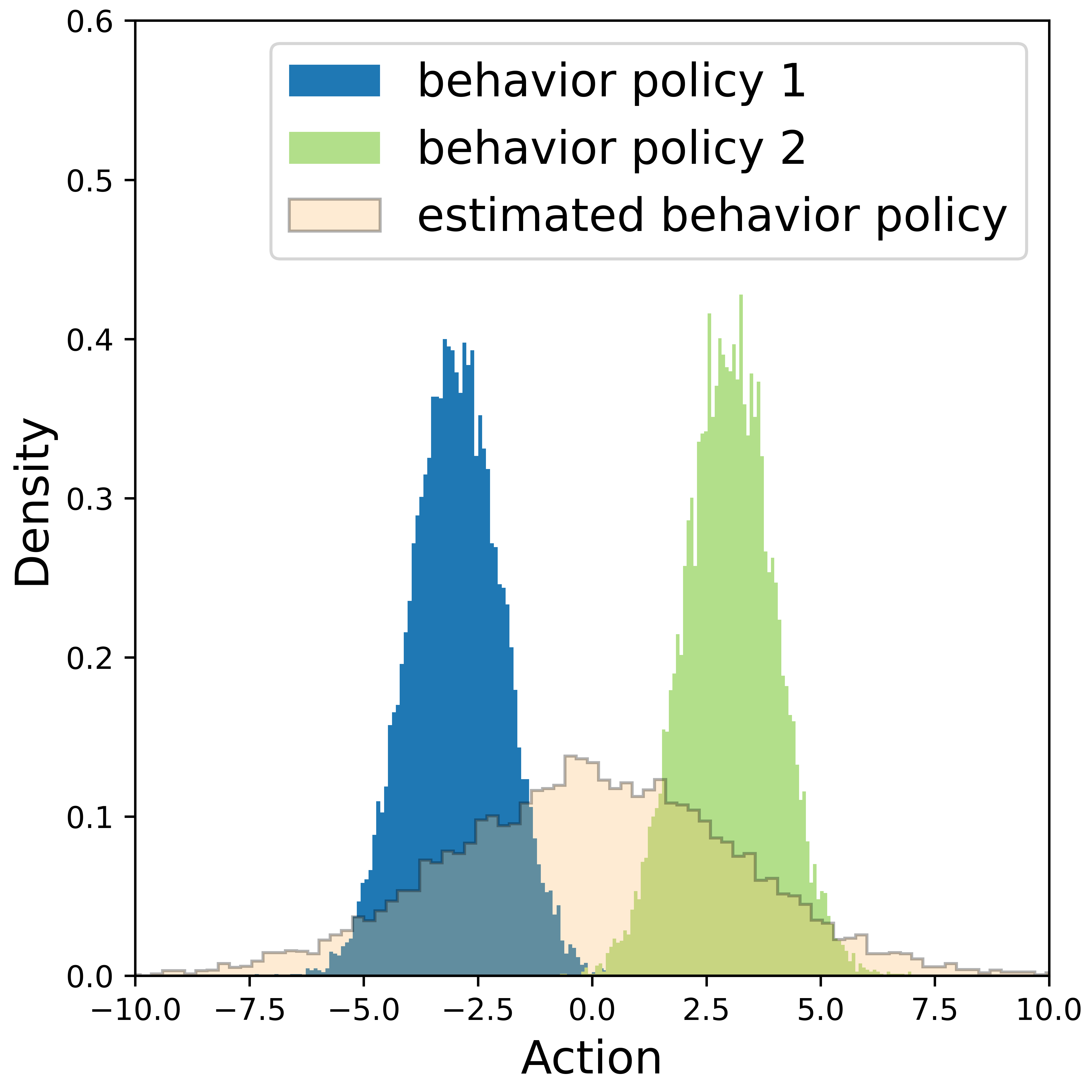}
     \end{subfigure}
     \caption{Example of multi-source trajectories (left) and behavior miss-specification (right). The green robot learns to play basketball using data collected from the blue and green demonstrators. The blue demonstrator prefers to pass the ball to the other agent (policy 1), while the green demonstrator prefers to shoot (policy 2). A single estimate for the behavior policy will place probability mass at regions not covered by data. }
     \label{fig:hetero_demo}
\end{figure}

In particular, the present study addresses behavior estimation on data collected from multiple sources, which is common when collecting demonstrations via crowdsourcing~\citep{pmlr-v87-mandlekar18a,mandlekar2021what}. Because different sources may prefer different actions, action distributions in such data can be multi-modal. \Cref{fig:hetero_demo} shows an minimalist example, where the green robot learns from data generated by two demonstrators. Because the blue and the green demonstrator prefers different actions, the empirical action distribution has two modes for that state. Now consider the canonical approach for behavior estimation that assumes a single Gaussian policy for continuous problems. As illustrated by the histogram in orange in~\Cref{fig:hetero_demo}, neglecting data heterogeneity, the canonical approach fails to model the data well.

Perhaps surprisingly, this canonical approach is widely adopted in literature~\citep{Gelada_Bellemare_2019,pmlr-v115-liu20a,pmlr-v119-pavse20a}, and data heterogeneity has not receive much attention. While learning latent action spaces using variational autoencoder (VAE)~\citep{DBLP:journals/corr/KingmaW13} seems to be a remedy~\citep{pmlr-v97-fujimoto19a,pmlr-v155-zhou21b}, the assumption of Gaussian action prior limits its expressiveness.

The present study claims that, though the empirical action distribution is multi-modal, each individual data source still induces uni-modal action distributions. Thus, data heterogeneity can be properly modeled by inferring a set of policies from data and the assignments of trajectories to these policies. In the example shown in Figure~\ref{fig:hetero_demo}, this corresponds to provide the agent with two policies and inferred assignments of trajectories. To this end, this work proposes a latent variable model and a model-learning algorithm to infer such a set from data. Furthermore, to demonstrate how the proposed model benefits offline RL, this work proposes localized BRAC-v, which integrates the proposed model into the BRAC-v algorithm. Since BRAC-v demonstrates reasonable performance on single-source data but degenerate on multi-source data~\citep{fu2020d4rl}, LBRAC-v showcases the efficacy of the proposed approach. 

For efficient model learning, the present study uses an embedding-based model parameterization and a vector-quantized variational inference algorithm. Policies and trajectories are embedded into a low-dimensional space, and their representation vectors are learned via action reconstruction. After model learning, agents can use these embeddings to retrieve the corresponding policies for trajectories.

This study validates its claims empirically using the D4RL benchmark~\citep{fu2020d4rl} and 15 new datasets. Experiment results show that algorithms that estimate single behavior policy worsened on multi-source data, which confirms the detriment of neglecting data heterogeneity. By contrast, LBRAC-v and existing VAE-based approaches achieved satisfactory performance. Moreover, LBRAC-v outperformed existing VAE-based approaches on datasets of moderate size and quality, while the existing approaches are better on large and high-quality datasets. These results illustrate the benefits and limitations of LBRAC-v. Lastly, with visualizations we show that the proposed model discovered patterns correlated with reward signals, which are further polished in policy learning. Increasing the size of the behavior set encourages clustering trajectories with behavior policies but inhibits such reward-related patterns. The contributions of this work are summarized as follows.
\begin{itemize}
    \item This work considers behavior estimation on multi-source data and highlights the detrimental effect of behavior misspecification.
    \item It proposes a latent variable model to learn a behavior set to overcome behavior miss-specification. In addition, it proposes LBRAC-v to showcase how the model benefits policy learning. 
    \item It demonstrates the efficacy of LBRAC-v with experiments and provides visualization-based explanations.
\end{itemize}

\section{Related Work}
\label{section:reference}
Prior art for offline RL discusses issues such as value overestimation~\citep{NEURIPS2020_0d2b2061} and poor convergence quality~\citep{Gelada_Bellemare_2019}, and practical algorithms often require behavior estimation for performing importance sampling~\citep{pmlr-v70-hallak17a,Gelada_Bellemare_2019,NEURIPS2019_cf9a242b} or behavior regularization~\citep{NEURIPS2019_c2073ffa,wu2019behavior,pmlr-v139-kostrikov21a}. However, none of them considers if data are from multiple sources. 

Meanwhile, it is worth mentioning the difference between this study and imitation learning from noisy demonstrations~\citep{pmlr-v97-wu19a,pmlr-v119-tangkaratt20a,pmlr-v139-wang21aa}. Although the latter opts for a well-performing policy, such a policy does not necessarily reproduce the data accurately. In consequence, quantities such as importance weights will be biased when computed using the policy.

Although several offline RL algorithms learn latent action spaces using VAE~\citep{pmlr-v97-fujimoto19a,pmlr-v155-zhou21b,Zhang_Shao_Jiang_He_Zhang_Ji_2022}, their assumption of Gaussian priors can be problematic. By contrast, the proposed model learns a discrete set of behavior policies that address multiple modality in data.

\section{Problem Statement}
\subsection{Preliminaries}
\paragraph{Markov Decision Process: } An infinite-horizon discounted Markov decision process (MDP) $\langle\mathcal{S}, \mathcal{A}, \mathcal{R}, P, \mathcal{\gamma}\rangle$ is a mathematical model for decision-making tasks. A state $s\in\mathcal{S}$ represents information provided to the agent for making a decision, and $\mathcal{S}$ is the set of all possible states. An action $a\in\mathcal{A}$ is an available option for a decision, and $\mathcal{A}$ is the set of possible actions. This study focuses on continuous control tasks, so actions are vectors in $\mathbb{R}^{d_\mathcal{A}}$ and $d_\mathcal{A}\in\mathbb{N}_+$. The reward function $R:\mathcal{S}\times\mathcal{A}\rightarrow \mathbb{R}$ provides the agent with scalar feedback for each decision. $P$ is a distribution over states conditioned on states and actions, which governs state transitions. $\gamma\in(0,1)$ is the discount factor used to define Q-functions.

\paragraph{Trajectories: } An MDP prescribes a sequential interaction procedure. Suppose the agent learns to navigate through a maze. Its current location is its states. At state $s\in\mathcal{S}$, it selects $a\in\mathcal{A}$ as the direction to go. Then, it is informed with its next state (next location) $s'\sim P(s|s,a)$ and reward $r=R(s, a)$. For example, $r=-1$ if the agent hits a wall, and $r=1$ if it reaches the target location. A trajectory is a sequence of states, actions and rewards. The $m$\textsuperscript{th} trajectory can be written as $\tau_m=(s_{m,1}, a_{m,1}, r_{m,1}, s_{m,2}, a_{m,2}, r_{m,2}, \dots)$, where the first and the second subscripts denote trajectory index and step index, respectively. In practice, trajectories are truncated to a finite length. The quality of $\tau_m$ is measured by its sum of discounted rewards $j(\tau_m)=\sum_{t=1}^\infty \gamma^{t-1}r_{m,t}$. A useful notion is transition. A transition from $\tau_m$ can be written as $(s_m,a_m,r_m,s'_m)$, which refers to two consecutive states, an action, and the corresponding reward for taking $a_m$ at $s_m$. 

\paragraph{Policies and Q-functions: }An agent selects actions with a policy $\pi:\mathcal{S}\rightarrow \Delta(\mathcal{A})$, where $\Delta(\mathcal{A})$ means the set of distributions over $\mathcal{A}$. The Q-function of $\pi$, $Q^\pi(s,a):\mathcal{S}\times\mathcal{A}\rightarrow\mathbb{R}$, gives the expected sum of discounted rewards obtained by taking $a\in\mathcal{A}$ at $s\in\mathcal{S}$ and following $\pi$ subsequently. That is, $Q^\pi(s) = \mathbb{E}_{\tau_m}\left[j(\tau_m)|s_{m,1}=s, a_{m,1}=a, \pi\right]$. $\pi$ is included in this expression to emphasize that it governs action selection at $t>1$. The objective of policy learning is to compute a policy $\pi^*$such that $Q^{\pi^*}(s,a)\geq Q^\pi(s,a)$, $\forall s,a\in\mathcal{S}\times\mathcal{A}$. 

\subsection{Behavior Estimation from Multi-source Data}
The input for behavior estimation is a fixed set of trajectories $\mathcal{D}=\{\tau_m\}_{m=1}^M$. Canonically, $\mathcal{D}$ is assumed to be generated by a single policy $b$ called the behavior policy. By contrast, this study considers $\mathcal{D}=\{\tau_m\}_{m=1}^M$ to be generated by multiple policies used by multiple sources. 

The output is a behavior set $\mathcal{B}=\{b_k\}_{k=1}^K$, where $b_k\in \mathcal{B}$ is a policy and $K\in\mathbb{N}_+$ is the number of policies. Besides, the learner also outputs an assignment matrix $G\in \{0,1\}^{M\times K}$ of trajectories to $\mathcal{B}$, which enables it to use $\mathcal{B}$ for policy learning. $G_{u,v}=1$ means that $\tau_u$ is considered to be generated by $b_v$, and $G_{u,v}=0$ otherwise, so $G$ allows for retrieving the corresponding $b_k\in \mathcal{B}$ for a given trajectory. To this end, the learner is provided with access to the index of trajectories, though for the sake of simplicity this index is only referenced in subscripts. 

The learning problem can be summarized as follows.
\begin{itemize}
   \item Input: offline trajectories $\mathcal{D}$.
   \item Output: $K$ behavioral policies $\mathcal{B}=\{b_k\}_{k=1}^K$ and the assignment matrix $G$.
\end{itemize}

\section{Learning the Behavior Set}
\begin{figure}[t]
    \centering
    \includegraphics[width=0.6\linewidth]{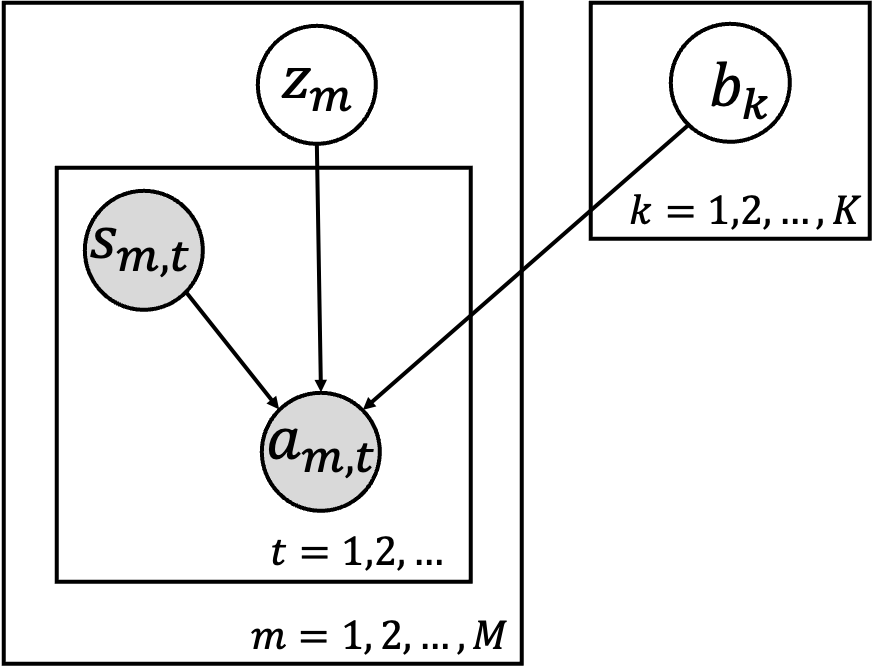}
    \caption{Graphical model representation for the proposed model. $m$ is the index of trajectories, and $t$ is the step index for states and actions in trajectories. $\{b_k\}_{k=1}^K$ is a set of $K$ behavior policies. To generate trajectory $\tau_m$, first sample a categorical variable $z_m$ supported on $\{1,2,\dots,K\}$. Then sample actions $a_{m,t}$ from $b_{z_m}(a|s_{m,t})$ once states $s_{m,t}$ are observed.}
    \label{fig:graphical_model}
\end{figure}

\begin{figure}[t]
    \centering
    \includegraphics[width=\linewidth]{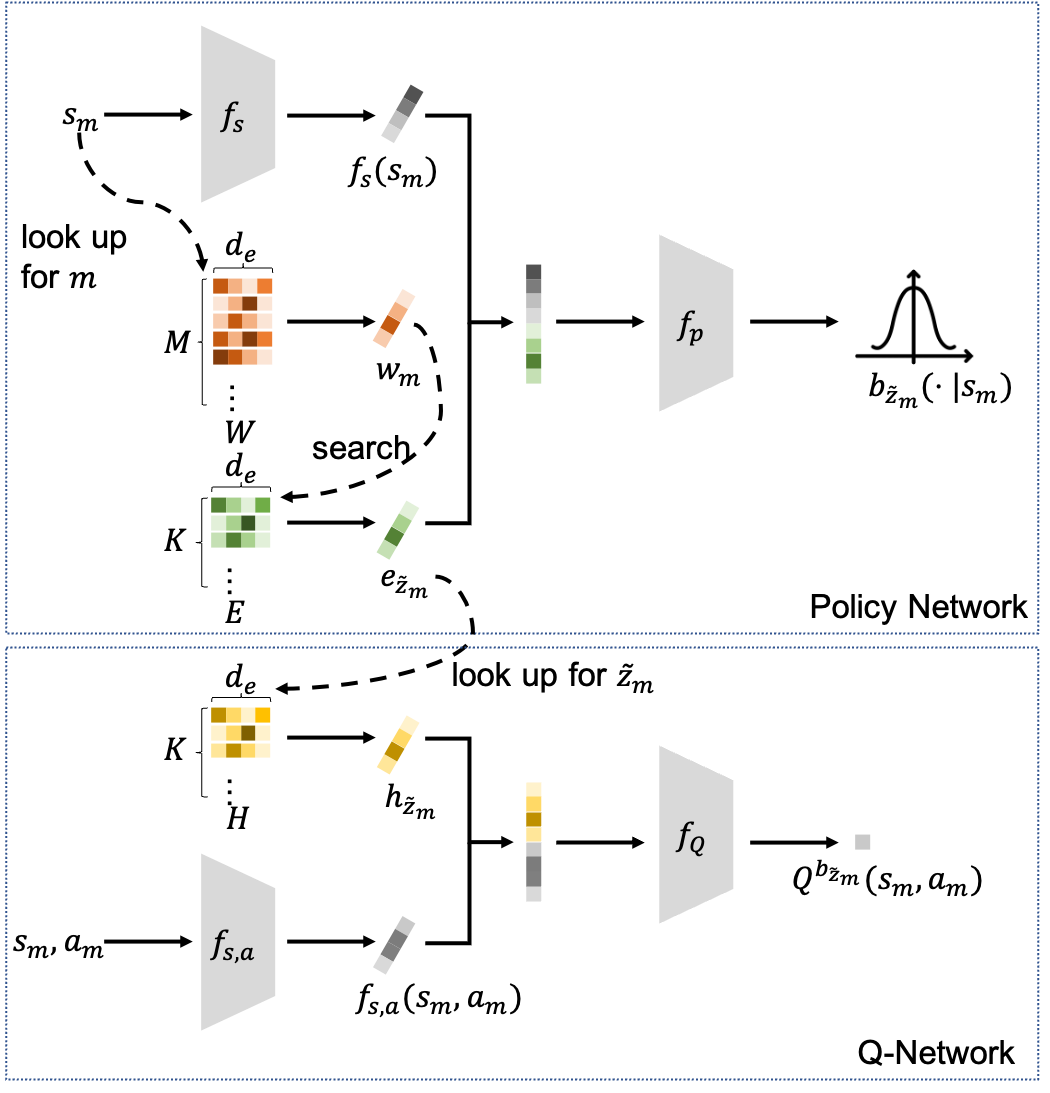}
    \caption{A diagram for the parameterization of the proposed model. Given $(s_m, a_m)$, the model looks up the $m$\textsuperscript{th} vector in $W$ and uses it to search for $e_{\tilde z_m}$ in $E$. See~\Cref{eq:posterior} for details of searching. $e_{\tilde z_m}$ is concatenated with vector $f_\phi(s_m)$ and passed to $f_\theta$ to generate actions. Meanwhile, the model looks up the $\tilde z_m$\textsuperscript{th} vector in $H$ and concatenates it with $g_\phi(s_m, a_m)$. The concatenated vector is passed to $g_\theta$ to compute Q values.}
    \label{fig:network}
\end{figure}

This section first describes the proposed latent variable model, its parameterization, and the model learning algorithm. Then, this section describes the proposed LBRAC-v algorithm for offline RL.
\subsection{Proposed Latent-Variable Model}
The proposed model assumes trajectories are generated by $K$ distinct policies. To generate trajectory $\tau_m$, we sample a categorical variable $z_m$ supported on $\{1,2,\dots,K\}$ and then roll out behavior policy $b_{z_m}$. This means that, at step $t$, we sample $a_{m,t}$ from $b_{z_m}(a|s_{m,t})$. Figure~\ref{fig:graphical_model} shows a graphical model representation for this process. 

A running assumption is that each trajectory is generated solely by one policy in $\mathcal{B}$. Accordingly, the proposed model exploits the identity of trajectories when learning $\mathcal{B}$. To understand the benefit, consider two transitions $(s_m, a_m, r_m, s'_m)$ and $(s_{m'}, a_{m'}, r_{m'}, s'_{m'})$, such that $s_m=s_{m'}$, $m\neq m'$, and $a_m\neq a_{m'}$. When $\tau_m$ and $\tau_{m'}$ are generated by different policies, $a_m$ and $a_{m'}$ are probably sampled from different Gaussian distributions. Existing VAE-based approaches becomes deficient in this case, as they assume a Gaussian prior for actions. However, by assigning $\tau_m$ and $\tau_{m'}$ to different elements in $\mathcal{B}$, the proposed model can leverage the simplicity of Gaussian policies without sacrificing expressiveness.

\Cref{fig:network} is a diagram for the proposed model. It consists of two major components: a policy network and a Q-network. The policy network represents $\mathcal{B}$ using an embedding matrix $E\in\mathbb{R}^{K\times d_e}$ and two functions $f_s$, and $f_p$, where $d_e\in\mathbb{N}_+$. Each row of $E$ is an embedding vector for a policy in $\mathcal{B}$. $f_s:\mathcal{S}\rightarrow \mathbb{R}^{d_\phi}$ is a function that encodes states, where $d_\phi\in\mathbb{N}_+$. $f_p:\mathbb{R}^{d_e}\times \mathbb{R}^{d_\phi}\rightarrow \mathbb{R}^{2\times d_\mathcal{A}}$  converts representations of states and policies to action distributions. For a state $s_m$, let $\mu_{s_m}$ and $\sigma_{s_m}$ be the mean and diagonal elements of the covariance matrix for the action distribution at $s_m$, respectively. Then,  
\begin{equation}
    [\mu_{s_m}^\intercal\ \sigma_{s_m}^\intercal] = f_p(e_{z_m}, f_s(s_m)).
    \label{eq:policy_repre}
\end{equation}

The Q-network of the proposed model uses an embedding matrix $H\in\mathbb{R}^{K\times d_e}$ and two functions $f_{s,a}$ and $f_Q$, to represent the Q-functions of policies in $\mathcal{B}$. Each row in $H$ is an embedding vector for the Q-function of a policy in $\mathcal{B}$. $f_{s,a}: \mathcal{S}\times\mathcal{A}\rightarrow \mathbb{R}^{d'_\phi}$ encodes state and actions in to vectors in $\mathbb{R}^{d'_\phi}$, where $d'_\phi\in\mathbb{N}_+$. $f_Q: \mathbb{R}^{d'_\phi+d_e}\rightarrow \mathbb{R}$ converts representations into Q-values:

\begin{equation}
    Q^{b_z} = f_Q(h_{z}, f_{s,a}(s, a)]).
    \label{eq:Q_repre}
\end{equation}

Note that in both the policy network and the Q-network, all policies (or Q-functions) share $f_s$ and $f_p$ (or $f_{s,a}$ and $f_Q$). This parameter-sharing mechanism reduces computational costs when compared to using $K$ separate policy networks (or Q functions). Moreover, it forces these functions to encode information that is common for all policies in $\mathcal{B}$, which is beneficial when using them for $\pi$.

\subsection{Parameter Learning}
We need to learn $E$, $H$, and parameters of $f_s$, $f_p$, $f_{s,a}$ and $f_Q$. \Cref{fig:network} depicts key operations of the learning algorithm. The main idea is to introduce a matrix $W\in\mathbb{R}^{M\times d_e}$ as variational parameters. Each row of this matrix is an embedding vector for a trajectory in $\mathcal{D}$. The proposed algorithm uses the following posterior distribution $q(z_m|s_m,a_m)$:
\begin{equation}
     q(z_m=k|s_m,a_m) =
    \begin{cases}
      1 & \text{for $k=\text{argmax}_{j}e_j^\intercal w_m$},\\
      0 & \text{otherwise}.
    \end{cases}       
    \label{eq:posterior}
\end{equation}
\noindent Let $\tilde z_m=\text{argmax}_{j}e_j^\intercal w_m$. This posterior distribution places all probability mass on $\tilde z_m$, and $e_{\tilde z_m}$ can be considered as the latent encoding for the behavior policy of $\tau_m$. 

Overall, parameters are learned via action reconstruction. Assuming a uniform prior for $z_m$, this in turn maximizes the evidence lowerbound for actions. However, care must be taken, as what relates $s_m$ and $a_m$ to $e_{\tilde z_m}$ in Equation~\ref{eq:posterior} is a non-differentiable searching operation. 

Inspired by~\citet{DBLP:conf/iclr/FortuinHLSR19}, the proposed algorithm uses an objective consists of three terms: $L_{\text{rec}}$, $L_{\text{com}}$, and $L_Q$. Specifically, $L_{\text{rec}}$ encourages action reconstruction. Recall that $f_p$ outputs action distributions, and let $\ell$ be the corresponding log-likelihood function. Then, $L_{\text{rec}}$ is defined as
\begin{equation}
    L_{\text{rec}} = -\ell(f_p(f_s(s_m), e_{\tilde z_m}); a_m) -\ell(f_p(f_s(s_m), w_m); a_m).
\label{eq:recon}
\end{equation}
\noindent The first term encourages $e_{\tilde z_m}$ to reconstruct actions, and the second term does likewise for $w_m$. Yet using~\Cref{eq:recon} alone, $w_m$ is not related to $e_{\tilde z_m}$, so $w_m$ and $e_{\tilde z_m}$ are not updated in a way that is consistent with $\tilde z_m=\text{argmax}_{j}e_j^\intercal w_m$. $L_{\text{com}}$ mitigates this issue by penalizing the discrepancy between them:
\begin{equation}
    L_{\text{com}} =1 - w_m^\intercal e_{\tilde z_m}.
\end{equation}

$L_Q$ is used for learning the Q-functions, which originates from fitted Q evaluation~\citep{pmlr-v97-le19a}.
\begin{equation}
    L_Q =\mathbb{E}_{a'\sim b_{\tilde z_m}} (r_m + \gamma \bar{Q}^{b_{\tilde z_m}}(s'_m, a') - Q^{b_{\tilde z_m}}(s_m, a_m))^2]
\end{equation}
\noindent $\bar{Q}^{b_{\tilde z_m}}$ is the target network for $Q^{b_{\tilde z_m}}$.

In summary, for a transition $(s_m, a_m, r_m, s'_m)$, the proposed algorithm utilizes the following objective function:
\begin{equation}
    L = L_{\text{rec}} + \alpha L_{\text{com}} + L_{\text{Q}},
    \label{eq:vae_obj}
\end{equation}
\noindent where $\alpha$ is a hyper-parameter fixed to $0.1$ in experiments. After learning, the assignment matrix $G$ can be computed using $W$ and $E$. The column index of the only non-zero element in the $m$\textsuperscript{th} row of $G$ is $\tilde z_m$. Additional details for the proposed model is provided in~\Cref{app:proposed_details}.

\subsection{The Proposed LBRAC-v Algorithm}
To demonstrate how the proposed model benefits offline RL, this study presents LBRAC-v, an extension of the BRAC-v algorithm~\citep{wu2019behavior}. To begin with, we first review BRAC-v and explains why it degenerates on multi-source data. 

BRAC-v leverages behavior regularization for offline RL. It penalizes the discrepancy between the policy being learned $\pi$ and the behavior policy $b$. On a transition, it minimizes the following objectives:

\begin{equation}
    \begin{split}
        L_{\mathrm{critic}}=\mathbb{E}_{\substack{a'\sim \pi(\cdot|s'_m)}} &\left[r_m+\gamma(\bar Q^\pi(s'_m,a')-\beta D(\pi, b, s'_m))\right.\\
        &\phantom{[}\left.-Q^\pi(s_m,a_m)\right]^2, \\
        L_{\mathrm{actor}} = \mathbb{E}_{a''\sim \pi(\cdot| s_m)}&\left[\beta D(\pi, b, s_m)-Q^\pi(s_m,a'')\right].
    \end{split}
    \label{eq:brac}
\end{equation}
\noindent $\bar Q^\pi$ is the target network for $Q^\pi$. $D(\pi, b, s)$ estimates the discrepancy between $\pi(\cdot|s)$ and $b(\cdot|s)$, and $\beta$ is a hyperparameter. Essentially, BRAC-v subtracts the Q value at $s'_m$ (or $s_m$) by $\beta D(\pi, b, s'_m)$ (or $\beta D(\pi, b, s_m)$) to penalize $\pi$ for deviating from $b$.~\citet{wu2019behavior} suggested using KL-divergence for $D(\pi, b, s)$:

\begin{equation}
    D(\pi, b, s) = \mathbb{E}_{a\sim \pi(\cdot|s)}\left[\log(\pi(a|s))-\log(b(a|s))\right].
\end{equation}

Because BRAC-v assumes a single behavior policy, it faces issues when handling multi-source data. Consider again the example in~\Cref{fig:hetero_demo}. As the mode of the estimated single behavior policy locates between modes of the two behavior policies, $D(\pi, b, s)$ unfortunately takes the minimum value between data modes. In consequence, BRAC-v encourages $\pi$ to take out-of-distribution actions.

To overcome this drawback, LBRAC-v integrates the proposed model into BRAC-v. When learning from $\tau_m$, it first find $b_{\tilde z_m}$ in $B$ and then penalizes the deviation of $\pi$ against $b_{\tilde z_m}$. This fine-grained characterization of behavior policies prevents $\pi$ to take out-of-distribution actions.

Another idea behind LBRAC-v is to reuse $f_s$, $f_p$, $f_{s,a}$, and $f_Q$ of a trained proposed model. Because these functions are adapted to the entire behavior set, they are also suitable for policies that are close to $\mathcal{B}$. Thus, LBRAC-v can benefit from the parameter-sharing mechanism of the proposed model. Specifically, LBRAC-v learns two vectors $e_\pi\in\mathbb{R}^{d_e}$ and $h_\pi\in\mathbb{R}^{d_e}$ as embedding vectors for $\pi$ and $Q^\pi$. $\pi$ is parameterized by replacing $e_{z_m}$ with $e_\pi$ in~\Cref{eq:policy_repre}, and $Q^\pi$ is parameterized by replacing $h_z$ with $h_\pi$ in~\Cref{eq:Q_repre}. In practice, $e_\pi$ and $h_\pi$ may be added to a trained proposed model to jointly optimize~\Cref{eq:brac} and~\Cref{eq:vae_obj}.

As a final remark, the proposed model can be straightforwardly integrated with other offline algorithms that utilizes parametric estimates of the behavior policy or latent action spaces. The present study presents LBRAC-v to illustrate how practitioners can extend existing algorithms for multi-source datasets.

\section{Experiments}
\label{section:experiment}

\begin{figure*}[ht!]
     \centering
     \begin{subfigure}[b]{0.3\textwidth}
         \centering
         \includegraphics[width=\textwidth]{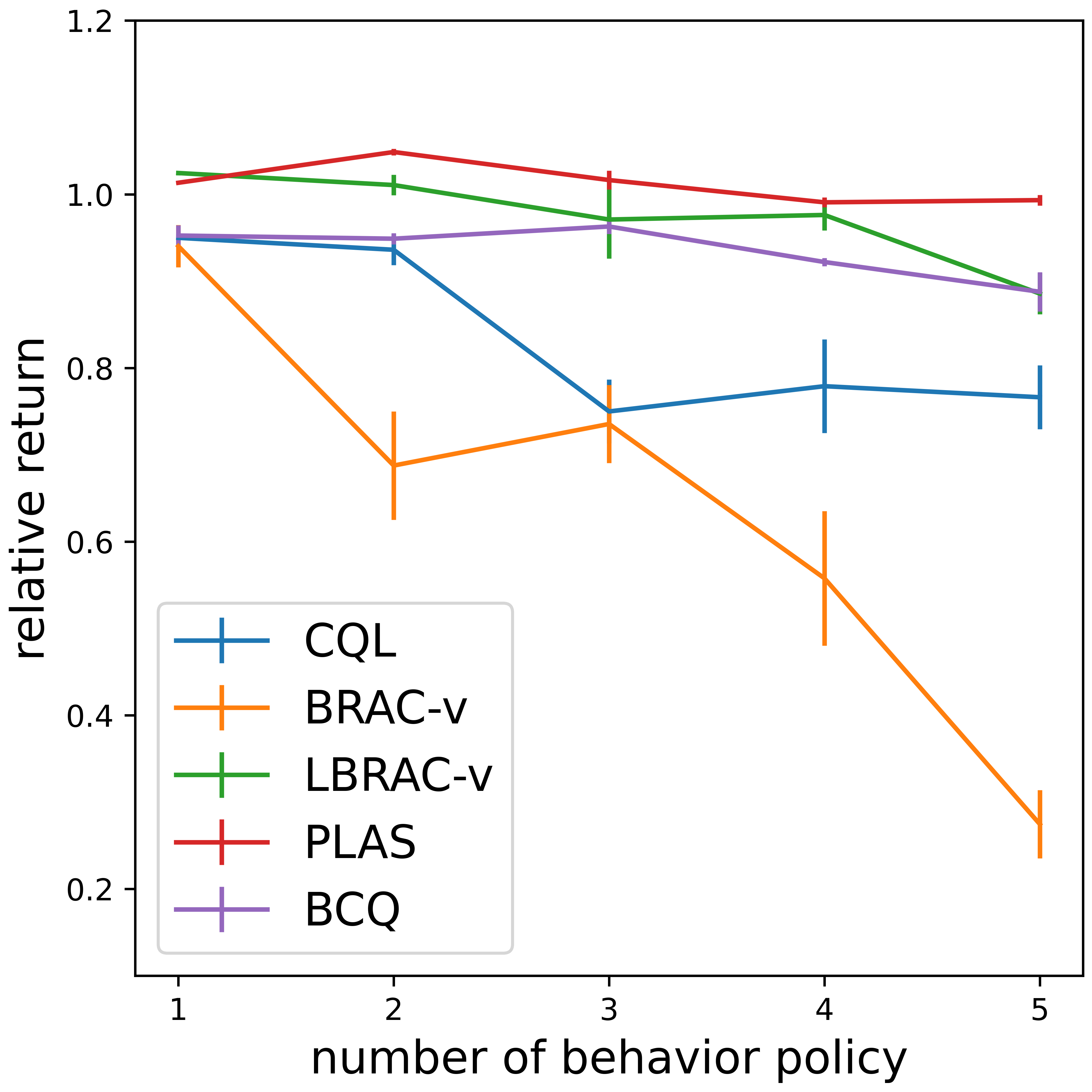}
         \caption{Halfcheetah.}
         \label{fig:halfcheetah_heterogeneous}
     \end{subfigure}
     \hfill
     \begin{subfigure}[b]{0.3\textwidth}
         \centering
         \includegraphics[width=\textwidth]{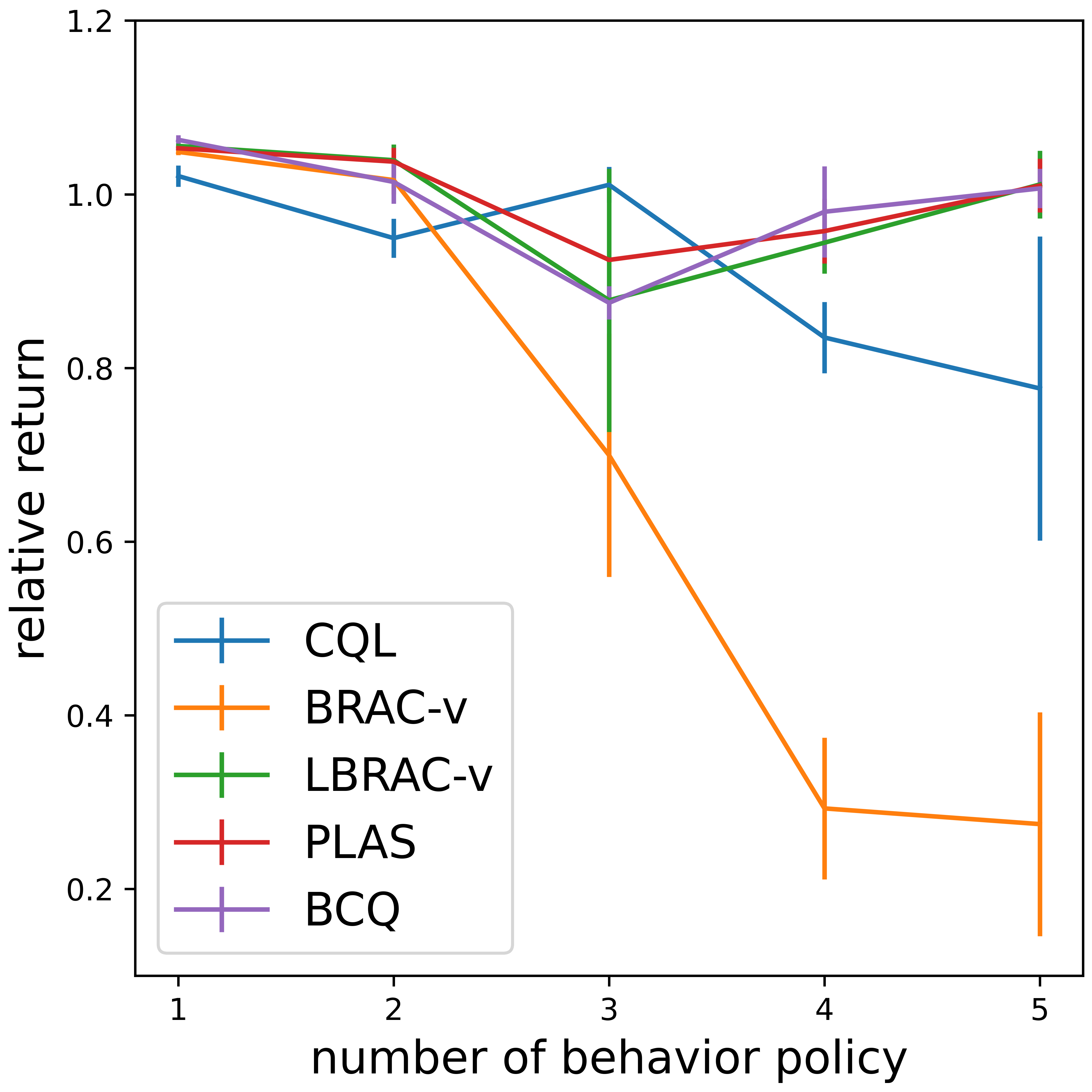}
         \caption{Walker2d.}
         \label{fig:walker2d_heterogeneous}
     \end{subfigure}
     \hfill
     \begin{subfigure}[b]{0.3\textwidth}
         \centering
         \includegraphics[width=\textwidth]{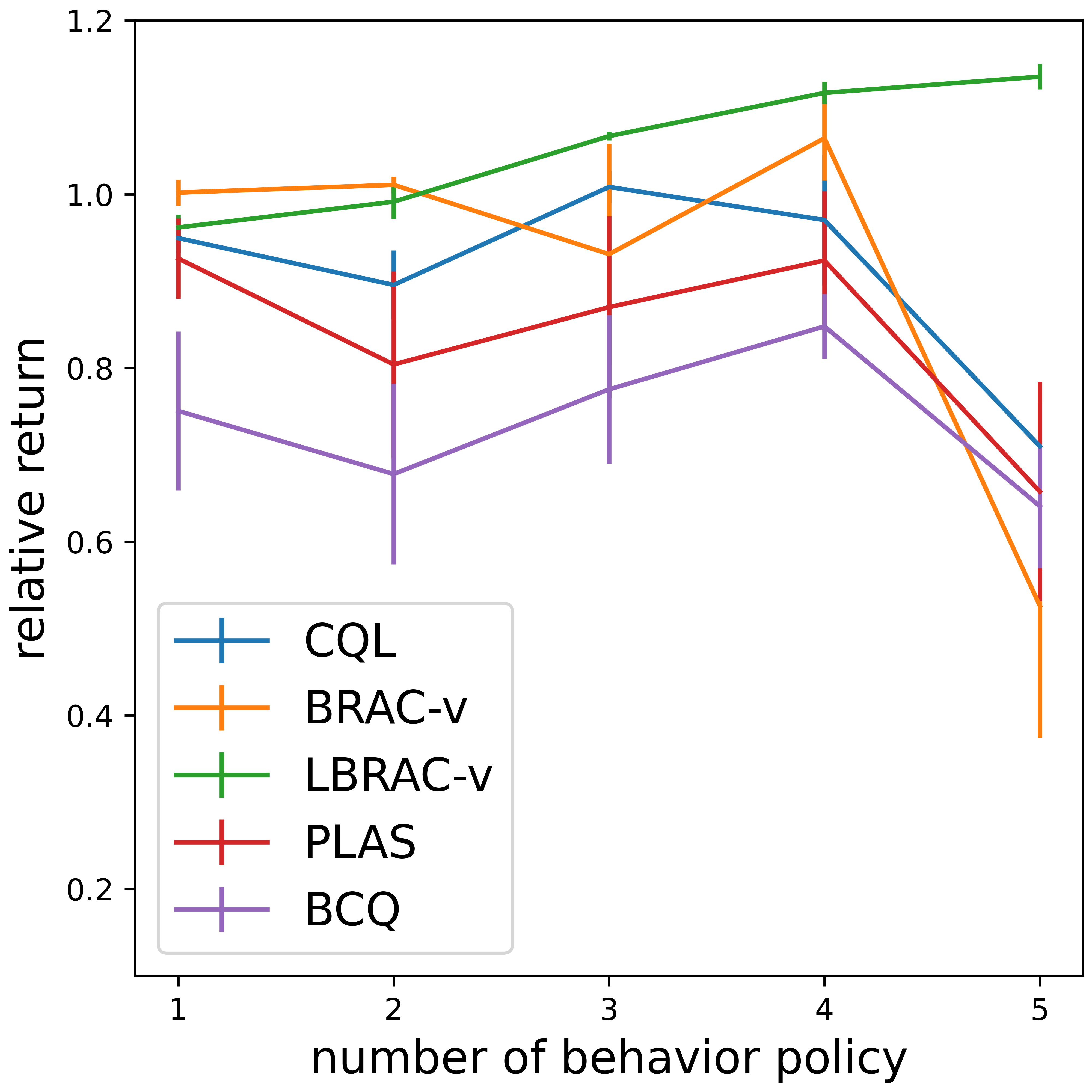}
         \caption{Hopper.}
         \label{fig:hopper_heterogeneous}
     \end{subfigure}
     \caption{Results on heterogeneous-$k$ datasets. The performance of CQL and BRAC-v degenerated when data were generated by multiple behavior policies, which confirms the existence of behavior miss-specification. The proposed LBRAC-v overcome such drawback of BRAC-v on multi-source data. VAE-based approaches were also robust on multi-source data.}
     \label{fig:heterogeneous_results}
\end{figure*}

\subsection{Datasets}

Three continuous control tasks were included in experiments: halfcheetah, walker2d, and hopper. For each task, four versions of datasets were taken from the D4RL benchmark~\citep{fu2020d4rl}. The random and medium versions were generated by a random policy and an RL agent, respectively. The medium-replay version contains transitions collected for training the RL agent, and the medium-expert version is a combination of medium version with expert demonstrations. The former two are single-source datasets, whereas the latter two are multi-source datasets.

However, as shown in~\Cref{tab:datasets} in~\Cref{app:exp_details}, for the same task these datasets differ in two of the three aspects: the number of behavior policies, the number of transitions, and the quality of behavior policies. They cannot directly reveal how the number of behavior policies affects performance. Thus, this study generates five new datasets for each task, denoted as heterogeneous-$k$ ($k\in\{1,2,3,4,5\}$)\footnote{These datasets are available at~\url{https://zenodo.org/record/7375417#.Y4Wzti9KGgQ}.}. For each task, we first trained five soft actor-critic (SAC)~\citep{Haarnoja2018} for an equal number of steps. Then, we generated the heterogeneous-$k$ version by rolling out the first $k$ agents for one million transitions. Each agent generated $1/k$ transitions. Consequently, heterogeneous-$k$ only differ in the number of behavior policies, which allows for investigating data heterogeneity.

Statistics of all the datasets and more details for heterogeneous-$k$ are provided in~\Cref{tab:datasets} in~\Cref{app:exp_details}.

\subsection{Alternative Methods}
Alternative methods are selected for three purposes. BC, BRAC-v, CQL, and MOPO are selected to demonstrate how data heterogeneity affects performance. MOPP and F-BRC are selected to show that latest algorithms with sophisticated parameterization for behavior policies still suffer from this issue. BCQ and PLAS are two existing VAE-based approaches, which are selected to investigate whether and when learning a behavior set is beneficial.
\begin{itemize}
     \item BC: a method that estimates single behavior policy.
     \item BRAC-v~\citep{wu2019behavior}: a model-free offline RL method that minimizes the KL divergence against an estimated single behavior policy. 
     \item CQL~\citep{NEURIPS2020_0d2b2061}: a model-free method that does not rely on behavior policies.
     \item MOPO~\citep{NEURIPS2020_a322852c}: a model-based algorithm that estimates state transition distributions.
     \item MOPP~\citep{ijcai2022p516}: a model-based algorithm that uses an ensemble of autoregressive dynamics models as the behavior policy.
     \item F-BRC~\citep{pmlr-v139-kostrikov21a}: a model-free method that minimizes the Fisher divergence against a mixture model for behavior policies.
     \item BCQ~\citep{pmlr-v97-fujimoto19a}: an algorithm that learns a VAE for actions.
     \item PLAS~\citep{pmlr-v155-zhou21b}: an algorithm that learns a VAE for actions and uses it to represent $\pi$.
\end{itemize}

Results on the D4RL datasets for BC, BRAC-v, CQL, and BCQ are taken from~\cite{fu2020d4rl}, and results for F-BRC, MOPO, MOPP, and PLAS are taken from the corresponding papers. For heteogeneous-$k$ datasets, the present study used the code provided by~\citet{fu2020d4rl} for BRAC-v and BCQ and the official code for CQL and PLAS.

\begin{table*}[ht]
\centering
\caption{Results on D4RL datasets. The best method for each dataset is highlighted in bold. LBRAC-v outperformed BRAC-v on eight datasets and achieved the best performance on four datasets. Compared to VAE-based methods (PLAS and BCQ), LBRAC-v performed better on medium and medium-replay versions.}
\scriptsize
\begin{tabular}{ccccccccccc}
\toprule
Task &Version   & BC & F-BRC &MOPP & PLAS  & BRAC-v    & CQL   & BCQ   &MOPO   & LBRAC-v                \\ \midrule
\multirow{4}{*}{halfcheetah} 
 &random        & 2.1 &33.3$\pm$1.3 & 9.4$\pm$2.6 & 25.8  & 31.2      & \textbf{35.4}  & 2.2   & \textbf{35.4$\pm$2.5}  & 26.37$\pm$0.29             \\
 &medium       & 36.1 & 41.3$\pm$0.3 & 44.7$\pm$2.6  & 42.2  & 46.3      & 44.4  & 40.7  & 42.3$\pm$1.6  & \textbf{52.14$\pm$0.11}   \\
 &medium-replay & 38.4 & 43.2$\pm$1.5 & 43.1$\pm$4.3  & 43.9  & 47.7      & 46.2  & 38.2  & \textbf{53.1$\pm$2.0}  & 48.41$\pm$0.14            \\
 &medium-expert & 35.8 & 93.3$\pm$10.2 & 106.2$\pm$5.1 & \textbf{96.6}  & 41.9      & 62.4  & 64.7  & 63.3$\pm$38.8 & 94.22$\pm$1.00            \\ \midrule
 
\multirow{4}{*}{walker2d} 
 &random        & 1.6 & 1.5$\pm$0.7 & 6.3$\pm$0.1 & 3.1   & 1.9       & 7.0   & 4.9   & \textbf{13.6$\pm$2.6}  & 11.96$\pm$4.39              \\
 &medium        & 6.6 & 78.8$\pm$1.0 & 80.7$\pm$1.0 & 66.9  & 81.1      & 79.2  & 53.1  & 17.8$\pm$19.3 & \textbf{82.40$\pm$1.30}   \\
 &medium-replay & 11.3 & 41.8$\pm$7.9 & 18.5$\pm$8.4  & 30.2  & 0.9       & 26.7  & 15.0  & 39.0$\pm$9.6  & \textbf{76.18$\pm$4.47}   \\
 &medium-expert & 6.4 & 105.2$\pm$3.9 & 92.9$\pm$14.1 & 89.6  & 81.6      & \textbf{111.0} & 57.5  & 44.6$\pm$12.9 & 109.65$\pm$0.23            \\ \midrule
\multirow{4}{*}{hopper} 
 &random        & 9.8 & 11.3$\pm$0.2 &\textbf{13.7$\pm$2.5} & 10.5  & 12.2      & 10.8  & 10.6  & 11.7$\pm$0.4  & 9.63$\pm$0.27             \\
 &medium        & 29.0 & 99.4$\pm$0.3 & 31.8$\pm$1.3 & 36.9  & 31.1      & 58.0  & 54.5  & 28.0$\pm$12.4 & \textbf{100.50$\pm$0.57}   \\
 &medium-replay & 11.8 & 35.6$\pm$1.0 & 32.3$\pm$5.9  & 27.9  & 0.6       & 48.6  & 33.1  & 67.5$\pm$24.7 & \textbf{81.89$\pm$12.77}  \\
 &medium-expert & 111.9 &\textbf{112.4$\pm$0.3} & 95.4$\pm$28 & 111.0 & 0.8       & 98.7  & 110.9 & 23.7$\pm$6.0  & 98.08$\pm$6.39           \\
 \bottomrule
\end{tabular}
\label{tab:performance}
\end{table*}

\subsection{Evaluation Metric}
Algorithms are compared by trajectory returns (i.e. non-discounted sum of rewards) obtained over 20 test runs. Returns are normalized into 0-100 scale as suggested by~\citet{fu2020d4rl}. The higher, the better. See~\Cref{app:exp_details} for more details.

Because of the inherent randomness of training RL agents, there is performance fluctuation among the behavior policies for heterogeneous-$k$. To eliminate this fluctuation, for experiments on these datasets the present study uses the ratio between normalized returns of an algorithm and the average normalized returns of behavior policies as evaluation metric. This metric is termed \textit{relative return}.

\subsection{Implementation Details}
$d_e$ was set to eight. $f_s$ and $f_p$ were parameterized by two layers of feed-forward networks with 200 hidden units, while $f_{s,a}$ and $f_Q$ were parameterized similarly but with 300 hidden units. The learning rates for the policy network and the Q-network were $5\times 10^{-5}$ and $1\times 10^{-4}$. Other details are available in~\Cref{app:proposed_details} and the code is available here\footnote{\url{https://github.com/Altriaex/multi_source_behavior_modeling}}. 

The proposed model and LBRAC-v are trained for $K\in\{1, 5, 10, 15, 20\}$. \Cref{tab:performance} reports the best value for each dataset. \Cref{tab:varying_k_d4rl} and~\cref{tab:varying_k_heterogeneous} report results for all $K$. Experiments were repeated for five different random seeds, and this paper reports the average value of metrics and standard deviations.

\subsection{Results}

\begin{figure*}[ht]
     \centering
     \includegraphics[width=0.85\textwidth]{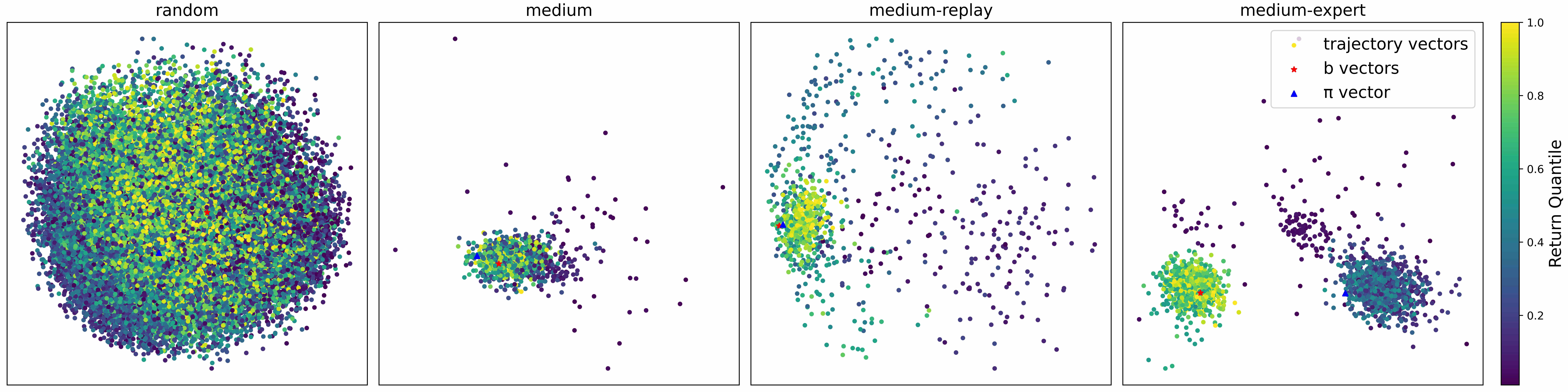}
     \caption{Visualizations of LBRAC-v on the four versions of walker2d datasets provided by D4RL. $K=1$. Trajectories were clustered by returns, especially for medium-replay and medium-expert versions.}
     \label{fig:visual_LBRAC_1}
\end{figure*}

\begin{figure*}[ht]
     \centering
     \includegraphics[width=0.85\textwidth]{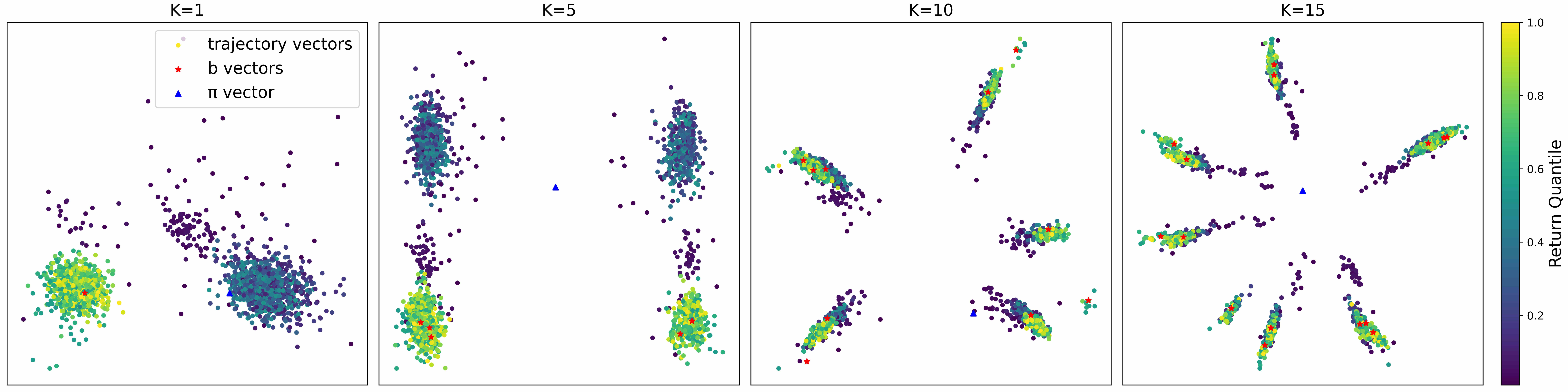}
     \caption{Visualizations of LBRAC-v on the medium-expert version of walker2d with different values of $K$. Increasing $K$ resulted in fine-grained clustering of trajectories around behavior policies, but reward-related patterns become less apparent.}
     \label{fig:visual_LBRAC_2}
\end{figure*}

\begin{figure}[ht]
     \centering
     \includegraphics[width=0.8\linewidth]{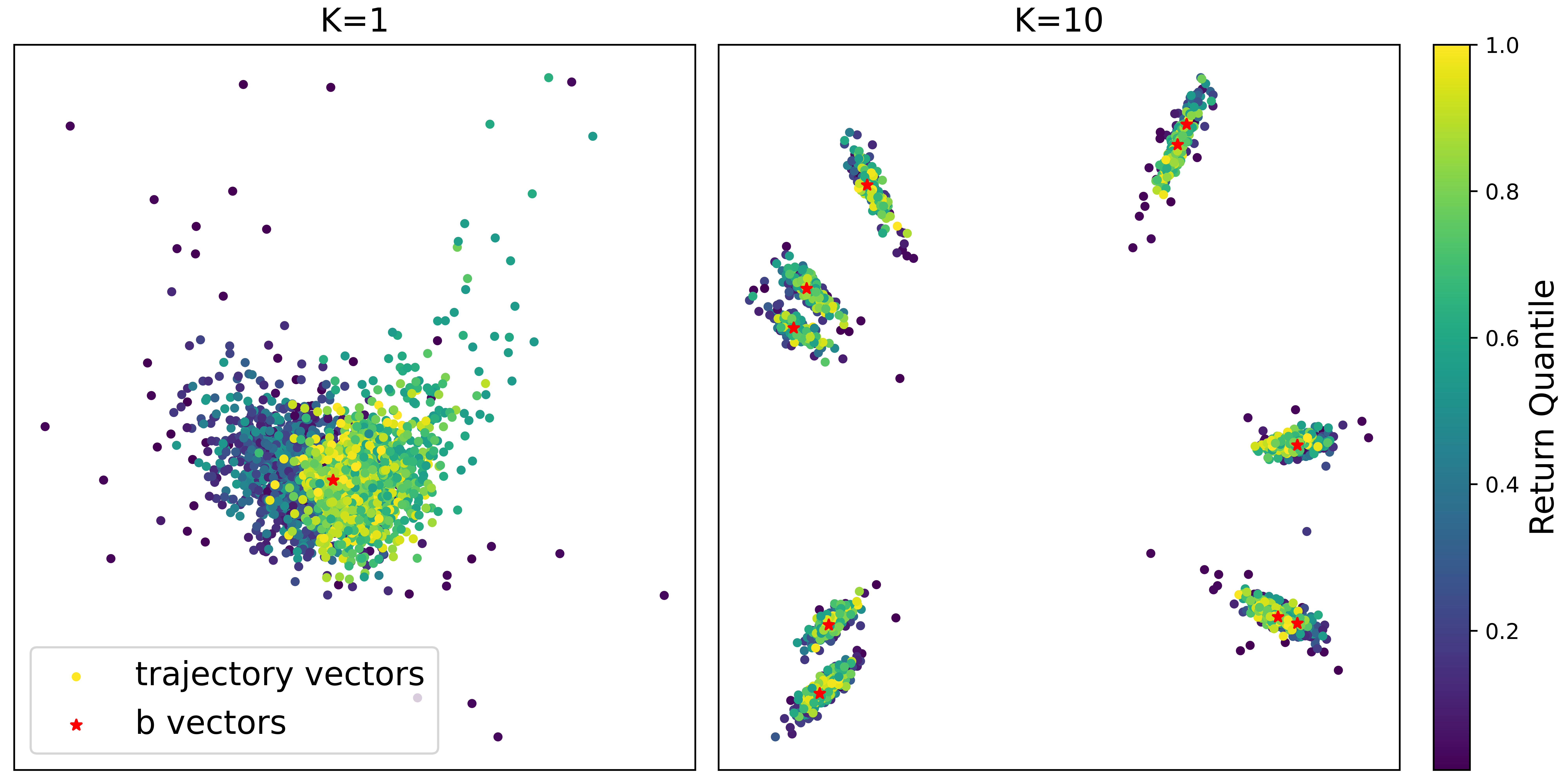}
     \caption{Visualizations of the proposed model on the medium-expert version of walker2d.}
     \label{fig:visual_proposed_model}
\end{figure}

First, compare LBRAC-v with BRAC-v, CQL, and MOPO. As shown in~\Cref{fig:heterogeneous_results}, performance of BRAC-v and CQL worsened on data generated by multiple policies. When $k=5$, BRAC-v lost about 50\% of the performance when $k=1$. In contrast, LBRAC-v outperformed BRAC-v and achieved consistent performance across $k$. \Cref{tab:performance} shows results on D4RL datasets. Although BRAC-v had moderate performance on medium versions, it failed on the medium-replay version of walker2d and the medium-replay and medium-expert version of hopper. CQL had strong performance on medium-expert versions but not on medium-replay versions. In contrast, LBRAC-v surpassed BRAC-v on 10 datasets and CQL on nine datasets. It also outperformed MOPO on medium versions and both multi-source versions, except for the medium-replay version of halfcheetah. These results confirm the existence of behavior miss-specification and the efficacy of LBRAC-v on both multi-source and single-source data.

Then, let us compare LBRAC-v with F-BRC and MOPP. Even though they use sophisticated parameterization for the behavior policy, they still assume a single behavior policy. \Cref{fig:heterogeneous_results} shows that LBRAC-v outperformed F-BRC on nine of the twelve datasets and MOPP on ten datasets. These results corroborate the importance of explicitly modeling multiple behavior policies from data.

Now compare LBRAC-v with PLAS and BCQ, the two VAE-based algorithms. As shown in~\Cref{fig:heterogeneous_results}, LBRAC-v outperformed BCQ and PLAS on the heterogeneous-$k$ versions for hopper, but it was outperformed by PLAS for halcheetah. For walker2d, the three methods had similar performance. These results indicate that both the VAE-based approach and the proposed model can model multi-source data. \Cref{tab:performance} reveals the strength and weakness of them. LBRAC-v outperformed PLAS on all the medium and medium-replay versions and the random version for halfcheetah and walker2d. It outperformed BCQ on every datasets except for the random and medium-expert versions of hopper. In short, PLAS and BCQ have better performance on the medium-expert versions but not for the rest. As shown in~\Cref{tab:datasets} in~\Cref{app:exp_details}, this version has one time more samples than other versions, and its average returns is also one time better. Thus, the VAE-based approach is suitable for large and high-quality datasets, whereas the proposed model is suitable for small or low-quality datasets.

Finally, this study provides insights about the proposals using visualizations. They were created by projecting $W$ to 2D space using principle component analysis. $E$ and $e_\pi$ were also projected to that space. \Cref{fig:visual_LBRAC_1} shows visualizations of LBRAC-v for walker2d datasets. $K$ is set to one to analyze effects of parts other than $E$. It shows that trajectories were clustered by returns, which indicates that sharing $w_m$ for actions in $\tau_m$ facilitates discovering reward-related patterns. For reference, the left part of~\Cref{fig:visual_proposed_model} shows the visualization of the proposed model on medium-expert version of walker2d when $K=1$. Reward-related patterns were also discovered, but they are less distinctive than those in~\Cref{fig:visual_LBRAC_1}. This observation implies that during policy learning return-related patterns are polished with information from rewards. 

\Cref{fig:visual_LBRAC_2} shows visualizations for various values of $K$. Increasing $K$, trajectory clusters were squeezed towards policy vectors, and reward-related patterns become less distinctive. The right part of~\Cref{fig:visual_proposed_model} reveals that it is the proposed model that failed to discover such patterns. Together, these visualizations suggest the followings.
\begin{enumerate}
    \item For small $K$, the proposed model discovers reward-related patterns for trajectories.
    \item Such patterns get polished in policy learning.
    \item Large value of $K$ encourages clustering trajectories by behavior policies but inhibits reward-related patterns.  
\end{enumerate}
Interested readers may find visualization obtained during training in~\Cref{fig:add_training} in~\Cref{app:additional_results} for more insights.

\section{Conclusion}
\label{section:conclusion}
Behavior estimation is a premise task of many offline RL algorithms. This work considered a scenario where training data are collected from multiple sources. We showed that it was detrimental to estimate a single behavior policy and overlook data heterogeneity in this case. To address this issue, this work proposed a latent variable model to estimate a set of behavior policies and integrated this model to the BRAC-v algorithm to showcase its usage. Empirical results confirmed the efficacy of the proposed model and the proposed extension. Moreover, visualizations showed that the proposed model discovered reward-related patterns for trajectories, which were further enhanced in policy learning. The present study is one of the few if any that addresses behavior estimation on multi-source data in literature and lays foundation for applying offline RL in real-world applications.

\section*{Acknowledgements}
This work was supported by JST CREST Grant Number JPMJCR21D1. The authors would like to thank Han Bao for valuable feedback on an early draft of this paper.

\bibliography{learning_behavioral_policy}

\begin{thebibliography}{28}
\providecommand{\natexlab}[1]{#1}

\bibitem[{Ba, Kiros, and Hinton(2016)}]{layernorm}
Ba, J.~L.; Kiros, J.~R.; and Hinton, G.~E. 2016.
\newblock Layer Normalization.
\newblock arXiv:1607.06450.

\bibitem[{Chen et~al.(2019)Chen, Beutel, Covington, Jain, Belletti, and
  Chi}]{10.1145/3289600.3290999}
Chen, M.; Beutel, A.; Covington, P.; Jain, S.; Belletti, F.; and Chi, E.~H.
  2019.
\newblock Top-K Off-Policy Correction for a REINFORCE Recommender System.
\newblock In \emph{Proceedings of the Twelfth ACM International Conference on
  Web Search and Data Mining}, 456--464. Melbourne, Australia: ACM.

\bibitem[{Dasari et~al.(2020)Dasari, Ebert, Tian, Nair, Bucher, Schmeckpeper,
  Singh, Levine, and Finn}]{pmlr-v100-dasari20a}
Dasari, S.; Ebert, F.; Tian, S.; Nair, S.; Bucher, B.; Schmeckpeper, K.; Singh,
  S.; Levine, S.; and Finn, C. 2020.
\newblock RoboNet: Large-Scale Multi-Robot Learning.
\newblock In \emph{Proceedings of the Third Conference on Robot Learning},
  885--897. Osaka, Japan: PMLR.

\bibitem[{Fortuin et~al.(2019)Fortuin, H{\"{u}}ser, Locatello, Strathmann, and
  R{\"{a}}tsch}]{DBLP:conf/iclr/FortuinHLSR19}
Fortuin, V.; H{\"{u}}ser, M.; Locatello, F.; Strathmann, H.; and R{\"{a}}tsch,
  G. 2019.
\newblock {SOM-VAE:} Interpretable Discrete Representation Learning on Time
  Series.
\newblock In \emph{Proceedings of the Seventh International Conference on
  Learning Representations}. New Orleans, LA, USA: OpenReview.net.

\bibitem[{Fu et~al.(2020)Fu, Kumar, Nachum, Tucker, and Levine}]{fu2020d4rl}
Fu, J.; Kumar, A.; Nachum, O.; Tucker, G.; and Levine, S. 2020.
\newblock D4RL: Datasets for Deep Data-Driven Reinforcement Learning.
\newblock arXiv:2004.07219.

\bibitem[{Fujimoto, Meger, and Precup(2019)}]{pmlr-v97-fujimoto19a}
Fujimoto, S.; Meger, D.; and Precup, D. 2019.
\newblock Off-Policy Deep Reinforcement Learning without Exploration.
\newblock In \emph{Proceedings of the Thirty-Sixth International Conference on
  Machine Learning}, 2052--2062. Long Beach, CA, USA: PMLR.

\bibitem[{Gelada and Bellemare(2019)}]{Gelada_Bellemare_2019}
Gelada, C.; and Bellemare, M.~G. 2019.
\newblock Off-Policy Deep Reinforcement Learning by Bootstrapping the Covariate
  Shift.
\newblock In \emph{Proceedings of the Thirty-Third AAAI Conference on
  Artificial Intelligence}, 3647--3655. Honolulu, Hawaii, USA: AAAI Press.

\bibitem[{Haarnoja et~al.(2018)Haarnoja, Zhou, Abbeel, and
  Levine}]{Haarnoja2018}
Haarnoja, T.; Zhou, A.; Abbeel, P.; and Levine, S. 2018.
\newblock Soft Actor-Critic: Off-Policy Maximum Entropy Deep Reinforcement
  Learning with a Stochastic Actor.
\newblock In \emph{Proceedings of the Thirty-Fifth International Conference on
  Machine Learning}, 1861--1870. Stockholm, Sweden: PMLR.

\bibitem[{Hallak and Mannor(2017)}]{pmlr-v70-hallak17a}
Hallak, A.; and Mannor, S. 2017.
\newblock Consistent On-Line Off-Policy Evaluation.
\newblock In \emph{Proceedings of the Thirty-Fourth International Conference on
  Machine Learning}, 1372--1383. Sydney, Australia: PMLR.

\bibitem[{Kingma and Welling(2014)}]{DBLP:journals/corr/KingmaW13}
Kingma, D.~P.; and Welling, M. 2014.
\newblock Auto-Encoding Variational Bayes.
\newblock In Bengio, Y.; and LeCun, Y., eds., \emph{Proceedings of the Second
  International Conference on Learning Representations}. Banff, AB, Canada.

\bibitem[{Kostrikov et~al.(2021)Kostrikov, Fergus, Tompson, and
  Nachum}]{pmlr-v139-kostrikov21a}
Kostrikov, I.; Fergus, R.; Tompson, J.; and Nachum, O. 2021.
\newblock Offline Reinforcement Learning with Fisher Divergence Critic
  Regularization.
\newblock In \emph{Proceedings of the Thirty-Eighth International Conference on
  Machine Learning}, 5774--5783. Virtual: PMLR.

\bibitem[{Kumar et~al.(2019)Kumar, Fu, Soh, Tucker, and
  Levine}]{NEURIPS2019_c2073ffa}
Kumar, A.; Fu, J.; Soh, M.; Tucker, G.; and Levine, S. 2019.
\newblock Stabilizing Off-Policy Q-Learning via Bootstrapping Error Reduction.
\newblock In \emph{Advances in Neural Information Processing Systems},
  volume~32, 11784--11794. Vancouver, BC, Canada: Curran Associates, Inc.

\bibitem[{Kumar et~al.(2020)Kumar, Zhou, Tucker, and
  Levine}]{NEURIPS2020_0d2b2061}
Kumar, A.; Zhou, A.; Tucker, G.; and Levine, S. 2020.
\newblock Conservative Q-Learning for Offline Reinforcement Learning.
\newblock In \emph{Advances in Neural Information Processing Systems},
  volume~33, 1179--1191. Virtual: Curran Associates, Inc.

\bibitem[{Lange, Gabel, and Riedmiller(2012)}]{Lange2012}
Lange, S.; Gabel, T.; and Riedmiller, M. 2012.
\newblock \emph{Batch Reinforcement Learning}, 45--73.
\newblock Springer Berlin Heidelberg.

\bibitem[{Le, Voloshin, and Yue(2019)}]{pmlr-v97-le19a}
Le, H.; Voloshin, C.; and Yue, Y. 2019.
\newblock Batch Policy Learning under Constraints.
\newblock In \emph{Proceedings of the Thirty-Sixth International Conference on
  Machine Learning}, 3703--3712. Long Beach, CA, USA: PMLR.

\bibitem[{Liu et~al.(2020)Liu, Swaminathan, Agarwal, and
  Brunskill}]{pmlr-v115-liu20a}
Liu, Y.; Swaminathan, A.; Agarwal, A.; and Brunskill, E. 2020.
\newblock Off-Policy Policy Gradient with Stationary Distribution Correction.
\newblock In \emph{Proceedings of the Thirty-Fifth Uncertainty in Artificial
  Intelligence Conference}, 1180--1190. Tel Aviv, Israel: PMLR.

\bibitem[{Mandlekar et~al.(2021)Mandlekar, Xu, Wong, Nasiriany, Wang, Kulkarni,
  Fei-Fei, Savarese, Zhu, and Mart{\'\i}n-Mart{\'\i}n}]{mandlekar2021what}
Mandlekar, A.; Xu, D.; Wong, J.; Nasiriany, S.; Wang, C.; Kulkarni, R.;
  Fei-Fei, L.; Savarese, S.; Zhu, Y.; and Mart{\'\i}n-Mart{\'\i}n, R. 2021.
\newblock What Matters in Learning from Offline Human Demonstrations for Robot
  Manipulation.
\newblock In \emph{Proceedings of the Fifth Annual Conference on Robot
  Learning}. London, UK.

\bibitem[{Mandlekar et~al.(2018)Mandlekar, Zhu, Garg, Booher, Spero, Tung, Gao,
  Emmons, Gupta, Orbay, Savarese, and Fei-Fei}]{pmlr-v87-mandlekar18a}
Mandlekar, A.; Zhu, Y.; Garg, A.; Booher, J.; Spero, M.; Tung, A.; Gao, J.;
  Emmons, J.; Gupta, A.; Orbay, E.; Savarese, S.; and Fei-Fei, L. 2018.
\newblock ROBOTURK: A Crowdsourcing Platform for Robotic Skill Learning through
  Imitation.
\newblock In \emph{Proceedings of the Second Conference on Robot Learning},
  879--893. Z{\"u}rich, Switzerland.

\bibitem[{Nachum et~al.(2019)Nachum, Chow, Dai, and Li}]{NEURIPS2019_cf9a242b}
Nachum, O.; Chow, Y.; Dai, B.; and Li, L. 2019.
\newblock DualDICE: Behavior-Agnostic Estimation of Discounted Stationary
  Distribution Corrections.
\newblock In \emph{Advances in Neural Information Processing Systems},
  volume~32. Vancouver, Canada: Curran Associates, Inc.

\bibitem[{Pavse et~al.(2020)Pavse, Durugkar, Hanna, and
  Stone}]{pmlr-v119-pavse20a}
Pavse, B.; Durugkar, I.; Hanna, J.; and Stone, P. 2020.
\newblock Reducing Sampling Error in Batch Temporal Difference Learning.
\newblock In \emph{Proceedings of the Thirty-Seventh International Conference
  on Machine Learning}, 7543--7552. Virtual.

\bibitem[{Tangkaratt et~al.(2020)Tangkaratt, Han, Khan, and
  Sugiyama}]{pmlr-v119-tangkaratt20a}
Tangkaratt, V.; Han, B.; Khan, M.~E.; and Sugiyama, M. 2020.
\newblock Variational Imitation Learning with Diverse-quality Demonstrations.
\newblock In \emph{Proceedings of the Thirty-Seventh International Conference
  on Machine Learning}, 9407--9417. Virtual: PMLR.

\bibitem[{Wang et~al.(2021)Wang, Xu, Du, and Lee}]{pmlr-v139-wang21aa}
Wang, Y.; Xu, C.; Du, B.; and Lee, H. 2021.
\newblock Learning to Weight Imperfect Demonstrations.
\newblock In \emph{Proceedings of the Thirty-Eighth International Conference on
  Machine Learning}, 10961--10970. Virtual: PMLR.

\bibitem[{Wu, Tucker, and Nachum(2019)}]{wu2019behavior}
Wu, Y.; Tucker, G.; and Nachum, O. 2019.
\newblock Behavior regularized offline reinforcement learning.
\newblock arXiv:1911.11361.

\bibitem[{Wu et~al.(2019)Wu, Charoenphakdee, Bao, Tangkaratt, and
  Sugiyama}]{pmlr-v97-wu19a}
Wu, Y.-H.; Charoenphakdee, N.; Bao, H.; Tangkaratt, V.; and Sugiyama, M. 2019.
\newblock Imitation Learning from Imperfect Demonstration.
\newblock In \emph{Proceedings of the Thirty-Sixth International Conference on
  Machine Learning}, Proceedings of Machine Learning Research, 6818--6827. Long
  Beach, CA, USA: PMLR.

\bibitem[{Yu et~al.(2020)Yu, Thomas, Yu, Ermon, Zou, Levine, Finn, and
  Ma}]{NEURIPS2020_a322852c}
Yu, T.; Thomas, G.; Yu, L.; Ermon, S.; Zou, J.~Y.; Levine, S.; Finn, C.; and
  Ma, T. 2020.
\newblock MOPO: Model-based Offline Policy Optimization.
\newblock In \emph{Advances in Neural Information Processing Systems},
  volume~33, 14129--14142. Virtual: Curran Associates, Inc.

\bibitem[{Zhan, Zhu, and Xu(2022)}]{ijcai2022p516}
Zhan, X.; Zhu, X.; and Xu, H. 2022.
\newblock Model-Based Offline Planning with Trajectory Pruning.
\newblock In \emph{Proceedings of the Thirty-First International Joint
  Conference on Artificial Intelligence}, 3716--3722. International Joint
  Conferences on Artificial Intelligence Organization.

\bibitem[{Zhang et~al.(2022)Zhang, Shao, Jiang, He, Zhang, and
  Ji}]{Zhang_Shao_Jiang_He_Zhang_Ji_2022}
Zhang, H.; Shao, J.; Jiang, Y.; He, S.; Zhang, G.; and Ji, X. 2022.
\newblock State Deviation Correction for Offline Reinforcement Learning.
\newblock In \emph{Proceedings of the Thirty-Sixth AAAI Conference on
  Artificial Intelligence}, 9022--9030. Virtual: AAAI Press.

\bibitem[{Zhou, Bajracharya, and Held(2020)}]{pmlr-v155-zhou21b}
Zhou, W.; Bajracharya, S.; and Held, D. 2020.
\newblock PLAS: Latent Action Space for Offline Reinforcement Learning.
\newblock In \emph{Proceedings of the Fourth Conference on Robot Learning},
  1719--1735. Virtual: PMLR.

\end{thebibliography}

\appendix
\clearpage

\section{Details for the Proposed Methods}\label{app:proposed_details}
This section presents details of the proposed model, the proposed learning algorithm and LBRAC-v. 

Except for $K$, this work uses same values of hyper-parameters on all datasets. The dimensionality of embedding spaces ($d_e$) is set to eight. $f_s$ is parameterized with two layers of feedforward networks whose hidden dimension and output dimension are 200 ($d_\phi=200$). The output of $f_s$ is processed with the layer norm operation~\citep{layernorm}. $f_p$ is parameterized with two layers of feedforward networks with 200 as hidden dimension and $2\times d_\mathcal{A}$ as output dimension. $f_Q$ and $f_{s,a}$ are parameterized similarly but with $d'_\phi=300$. This work uses the Relu function as the activation function for all hidden layers and output of $f_s$ and $f_{s,a}$. 

The output of $f_p$ parameterizes an action distribution as follows. $f_p$ uses the location-scale reparameterization to compute Gaussian action distributions using $\mathcal{N}(0,I_{d_\mathcal{A}})$, where $I_{d_\mathcal{A}}$ means an identity matrix in $\mathbb{R}^{d_\mathcal{A}}$. The mean part of action distribution, $\mu_{s}$, is first squashed into $[-1,1]$ using the $\tanh(\cdot)$ function and then mapped to the range of actions via affine transformation. $f_p$ in fact outputs $\log(\sigma_s)$ instead of $\sigma_s$ directly, which is clipped to $[-10, 10]$ before passing to the exponential function. Then a multi-variate Gaussian distribution for actions can be obtained with $\mu_s$ and $\sigma_s$.

Meanwhile, similar to BRAC-v, the present study utilizes an ensemble of two functions for the Q network part. Each of them contains a set of $f_{s,a}$, $H$ and $f_Q$.  

\Cref{algo:proposed_model} presents pseudocode for learning the proposed model. This work uses $5\times 10^{-5}$ as the learning rate for $f_s$, $f_p$, $W$ and $E$. For $f_{s,a}$, $H$ and $f_Q$, the learning rate is set to $1\times 10^{-4}$. The rate of soft update for target networks is $0.001$. Rows of $E$, $W$, and $H$ are normalized by $l_2$-norm. 

\Cref{algo:lbrac} presents pseudocode for the proposed LBRAC-v. All of the hyper-parameters are the same with the learning algorithm for the proposed model. Before policy learning, $e_\pi$ and $h_\pi$ is initialized with the vector of the best policy in $\mathcal{B}$ and its Q-function. The best policy is determined as follows. First compute the percentiles $\{q_i\}$ of trajectories returns and use them to discretize returns. In other words, the discretized return of $\tau_m$, $g_m$, is $j$ if $q_j \leq \sum_{t}r_{m,t} < q_{j+1}$, where $q_j$ and $q_{j+1}$ are two consecutive percentile values such as 95 and 96. This quantization is employed to improve the robustness against extreme values. Then for each $b_k\in\mathcal{B}$, compute the average of discretized returns of the trajectories assigned to it. The one with the largest average discretized return is considered as the best behavior policy.

\begin{algorithm}[ht]
    \caption{Learning the proposed model}
    \label{algo:proposed_model}
    \KwIn{$D=\{(s_m, a_m, r_m, s'_m)\}_{m=1}^M$\newline
    $T$, the number of training steps}
    \KwOut{$\{b_k \}_{k=1}^K$ and $\{Q^{b_k}\}_{k=1}^K$}
    Initialize $f_s$, $f_p$, $f_{s,a}$, $f_Q$, $W$, $E$ and $H$. \\
    \For{$t=1,2,\dots,T$}{
        Sample a batch of transitions $D_{\mathrm{batch}}$ from $D$.\\
        Look up $w_m$ for transitions in $D_{\mathrm{batch}}$ and normalize them by $\ell_2$ norm.\\
        Normalize rows of $E$ by $\ell_2$ norm, and compute $\tilde z_m=\mathrm{argmax}_j e_j w_m$ for $w_m$ of $D_{\mathrm{batch}}$\\
        Update $f_s$, $f_p$, $W$ and $E$ by minimizing $L_\mathrm{rec}+\alpha L_\mathrm{com}$.\\  
        Normalize rows of $H$ by $\ell_2$ norm, and update $f_{s,a}$, $f_Q$, $H$ by minimizing $L_Q$. \\
    }
\end{algorithm}

\begin{algorithm}[ht]
    \caption{LBRAC-v}
    \label{algo:lbrac}
    \KwIn{$D=\{(s_m, a_m, r_m, s'_m)\}_{m=1}^M$,\newline
    $T$, the number of training steps\newline
    Trained $f_s$, $f_p$, $f_{s,a}$, $f_Q$, $W$, $E$ and $H$.}
    \KwOut{$\pi$}
    Compute the percentiles $\{q_j\}_{j=0}^{99}$ of $\{\sum_{t}r_{m,t}\}_{m=1}^M$.\\
    Compute the discretized returns $g_m$. $g_m=j$ if $q_j \leq \sum_{t}r_{m,t} < q_{j+1}$.\\ 
    For each $b_k$, compute $u_k= \frac{\sum_{m=1}^M G_{k,m} * \bar g_m}{\sum_{m=1}^M G_{k,m}}$ \\
    Let $k^*=\mathrm{argmax}_k u_k$, and initialize $e_\pi$ as $e_{k^*}$ and the  $h_\pi$ as $h_{k^*}$. \\
    \For{$t=1,2,\dots,T$}{
        Sample a batch of transitions $D_{\mathrm{batch}}$ from $D$.\\
        Look up $w_m$ for transitions in $D_{\mathrm{batch}}$ and normalize them by $\ell_2$ norm.\\
        Normalize rows of $E$ by $\ell_2$ norm, and compute $\tilde z_m=\mathrm{argmax}_j e_j w_m$ for $w_m$ of $D_{\mathrm{batch}}$\\
        Update $f_s$, $f_p$, $W$, $E$ and $e_\pi$ by minimizing $L_\mathrm{rec}+\alpha L_\mathrm{com}+L_\mathrm{actor}$.\\  
        Normalize rows of $H$ by $\ell_2$ norm, and update $f_{s,a}$, $f_Q$, $H$ and $h_\pi$ by minimizing $L_Q+L_\mathrm{critic}$. \\
    }
\end{algorithm}

\section{Details for Experiments}\label{app:exp_details}

\subsection{Details for heterogeneous-$k$ Datasets}
The proposed heterogeneous-$k$ datasets are generated in a similar way as the medium versions in the D4RL benchmark. The behavior policies are trained using the SAC algorithm implemented in rlkit repository\footnote{https://github.com/rail-berkeley/rlkit}. As in practice dataset provided to offline policy learners are rarely of superior quality, the behavior policies are trained from scratch for only one million gradient steps, while by default SAC agents are trained for three million gradient steps. The five behavior policies for a task are trained with random seeds one to five.~\Cref{tab:heterogeneous_behavior_returns} shows the normalized returns of these behavior policies.

\begin{table}[ht]
\caption{Normalized returns of behavior policies of heterogeneous-$k$ datasets.}
\label{tab:heterogeneous_behavior_returns}
\scriptsize
\begin{tabular}{@{}llllll@{}}
\toprule
            & 1\textsuperscript{st} policy & 2\textsuperscript{nd} policy & 3\textsuperscript{rd} policy & 4\textsuperscript{th} policy & 5\textsuperscript{th} policy \\\midrule
halfcheetah & 72.75      & 81.90      & 84.78      & 84.92      & 66.86      \\
waker2d     & 88.43      & 103.84     & 85.32      & 100.91     & 83.10            \\
hopper      & 108.57     & 106.03     & 99.69      & 79.87      & 85.33            \\ \bottomrule
\end{tabular}
\end{table}

For each task, the heterogeneous-$k$ dataset is generated by rolling out the first $k$ policies using the code provided by~\citet{fu2020d4rl}. Each policy generates an even amount of transitions. Following the standard of these tasks, a trajectory starts after reseting the environment, and it ends if a terminal signal is received, or the number of steps reaches 1000. Statistics of these datasets can be found in~\Cref{tab:datasets}.   

\begin{table}[t]
\caption{Statistics of datasets. The random, medium, medium-replay, and medium-expert versions are provided by~\citet{fu2020d4rl}. The heterogeneous-$k$ ($k\in\{1,2,3,4,5\}$) versions are proposed by the present study. The return column shows the average of normalized trajectory return in a dataset.}
\label{tab:datasets}
\centering
\scriptsize
\begin{tabular}{ccrrr}
\toprule
Task    &Version            & M     & \#transitions & Normalized Return \\ \midrule
\multirow{4}{*}{halfcheetah} 
        & random            & 1000  & 1000000       & -0.26             \\
        & medium            & 1000  & 1000000       & 155.8             \\
        & medium-replay     & 202   & 202000        & 103.65            \\
        & medium-expert     & 2000  & 2000000       & 245.61            \\
        & heterogeneous-1   & 1000  & 1000000       & 69.91             \\
        & heterogeneous-2   & 1000  & 1000000       & 73.91             \\
        & heterogeneous-3   & 1000  & 1000000       & 76.14             \\
        & heterogeneous-4   & 1000  & 1000000       & 77.47             \\
        & heterogeneous-5   & 1000  & 1000000       & 74.76             \\ \midrule
\multirow{4}{*}{walker2d} 
        & random            & 48907 & 1000000       & 0.01              \\
        & medium            & 513   & 1000000       & 202.84            \\
        & medium-replay     & 791   & 302000        & 28.90             \\
        & medium-expert     & 514   & 1999995       & 496.24            \\
        & heterogeneous-1   & 1037  & 1000000       & 88.79             \\
        & heterogeneous-2   & 1015  & 1000000       & 88.62             \\
        & heterogeneous-3   & 1018  & 1000000       & 85.93             \\
        & heterogeneous-4   & 1006  & 1000000       & 92.33             \\
        & heterogeneous-5   & 1018  & 1000000       & 92.29             \\ \midrule
\multirow{4}{*}{hopper} 
        & random            & 45239 & 1000000       & 1.19              \\
        & medium            & 2185  & 1000000       & 44.32             \\
        & medium-replay     & 1639  & 402000        & 18.49             \\
        & medium-expert     & 2274  & 1999906       & 91.33             \\
        & heterogeneous-1   & 1180  & 1000000       & 94.72             \\
        & heterogeneous-2   & 1207  & 1000000       & 93.23             \\
        & heterogeneous-3   & 1454  & 1000000       & 76.67             \\
        & heterogeneous-4   & 1513  & 1000000       & 73.08             \\
        & heterogeneous-5   & 1695  & 1000000       & 65.88             \\ 
\bottomrule
\end{tabular}
\end{table}

\subsection{Details for Other Datasets}
Datasets other than the heterogeneous-$k$ are released by~\citet{fu2020d4rl}. As shown in~\Cref{tab:datasets}, for the same tasks they differ in behavior policies, size, trajectory returns and number of transitions. In particular,~\Cref{fig:medium_replay_return} shows the distribution of return for the medium-replay versions. It shows that the return distribution for halfcheetah appears to be bi-modal, containing many high-return trajectories. In contrast, most of the trajectories in this version for walker2d and hopper are low-return trajectories.    

\subsection{Return Normalization}
Throughout this paper, trajectory returns are normalized as suggested by~\citet{fu2020d4rl}. The test returns of an algorithm obtained for some task is transformed using the equation $100 * \frac{\mathrm{return}-\mathrm{random}\ \mathrm{return}}{\mathrm{expert}\ \mathrm{return}-\mathrm{random}\ \mathrm{return}}$. This transformation eliminates the range difference in returns and makes results more comprehensible. Values of random returns and expert returns are taken from~\citep{fu2020d4rl}. In specific, the return for a random policy on halfcheetah, walker2d and hopper are -280.18, 1.63 and -20.27, respectively. The return for expert demonstrations on these tasks are 12135.0, 4592.3, and 3234.3.

\subsection{Details for Alternative Methods}
For D4RL datasets, results of BRAC-v, CQL and BCQ are taken from~\cite{fu2020d4rl}, and results of PLAS and MOPO are taken from their original paper. For the proposed datasets, the present study runs the code provided by~\citet{fu2020d4rl} for BRAC-v and BCQ, the code released in their original papers for CQL, PLAS and MOPO. However, some modifications is required for adapt the code for CQL for the latest version of PyTorch platform.

Following the methodology of~\citet{fu2020d4rl}, algorithms are trained for 0.5 million gradient steps. For other, the present study sticks with the value suggested in the corresponding code base or papers.

\section{Additional Results}\label{app:additional_results}
\subsection{Discussion for Medium-Replay Datasets}
\begin{figure*}[t]
     \centering
     \begin{subfigure}[b]{0.31\textwidth}
         \centering
         \includegraphics[width=\textwidth]{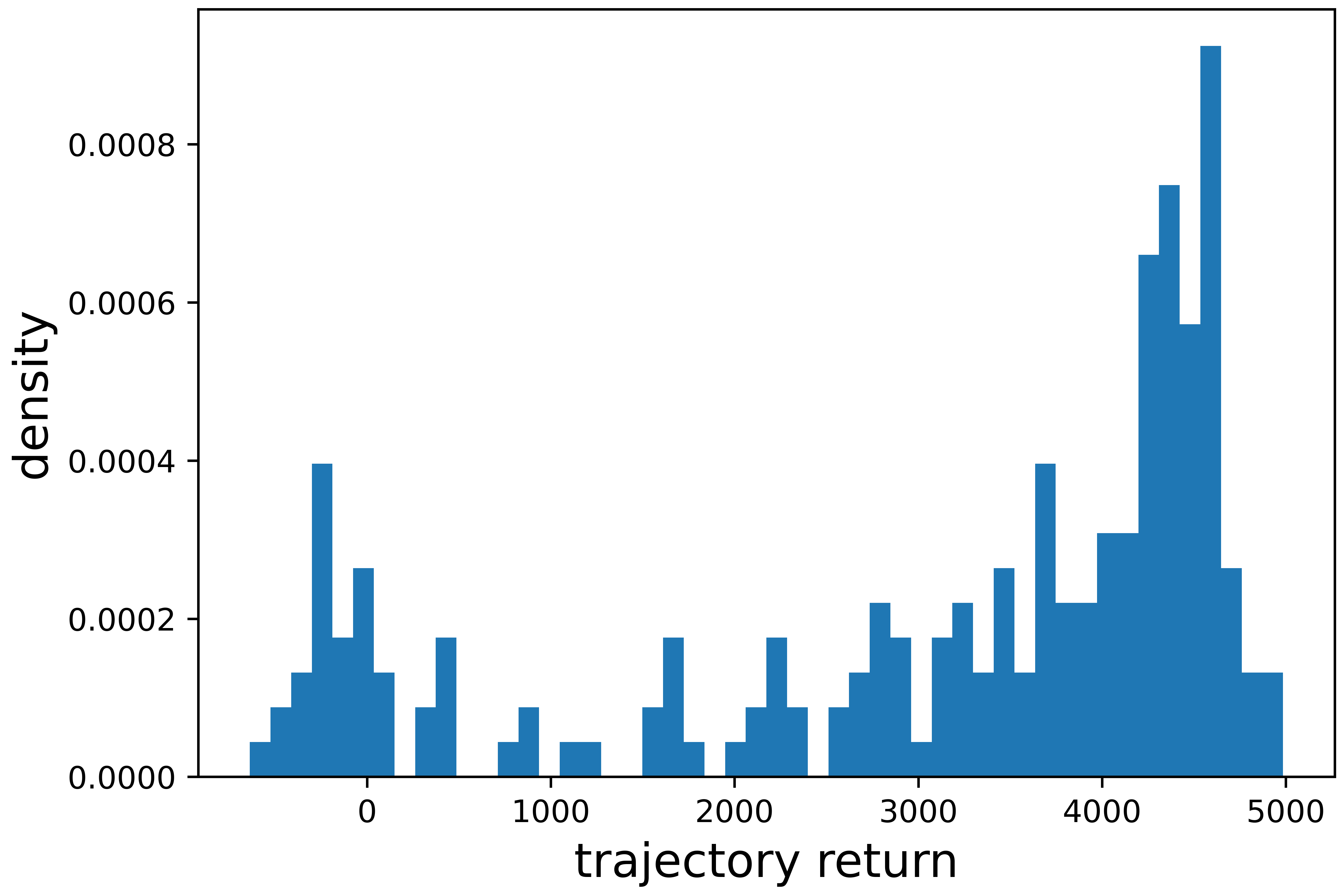}
         \caption{Halfcheetah}
     \end{subfigure}
     \hfill
     \begin{subfigure}[b]{0.31\textwidth}
         \centering
         \includegraphics[width=\textwidth]{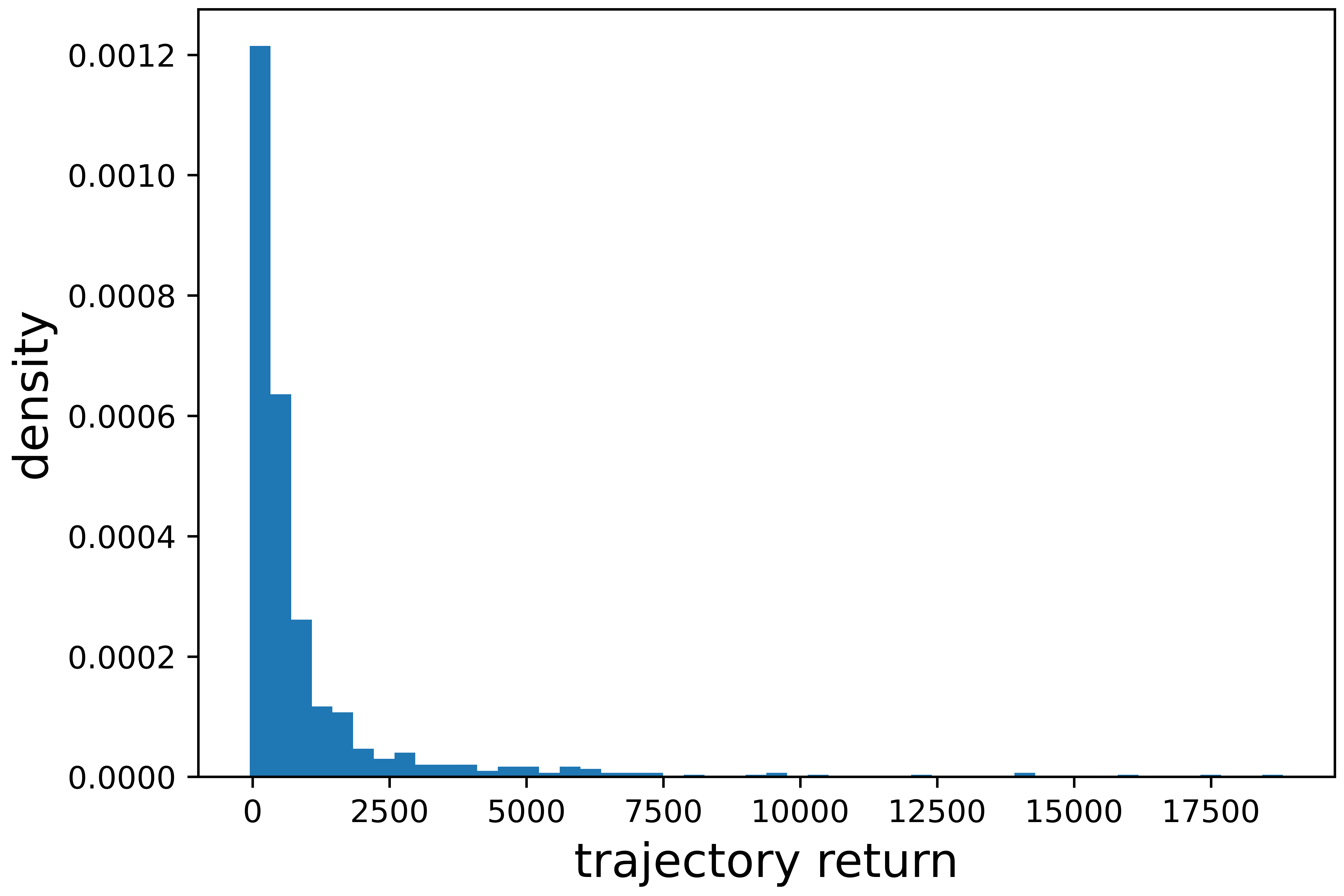}
         \caption{Walker2d}
     \end{subfigure}
     \hfill
     \begin{subfigure}[b]{0.31\textwidth}
         \centering
         \includegraphics[width=\textwidth]{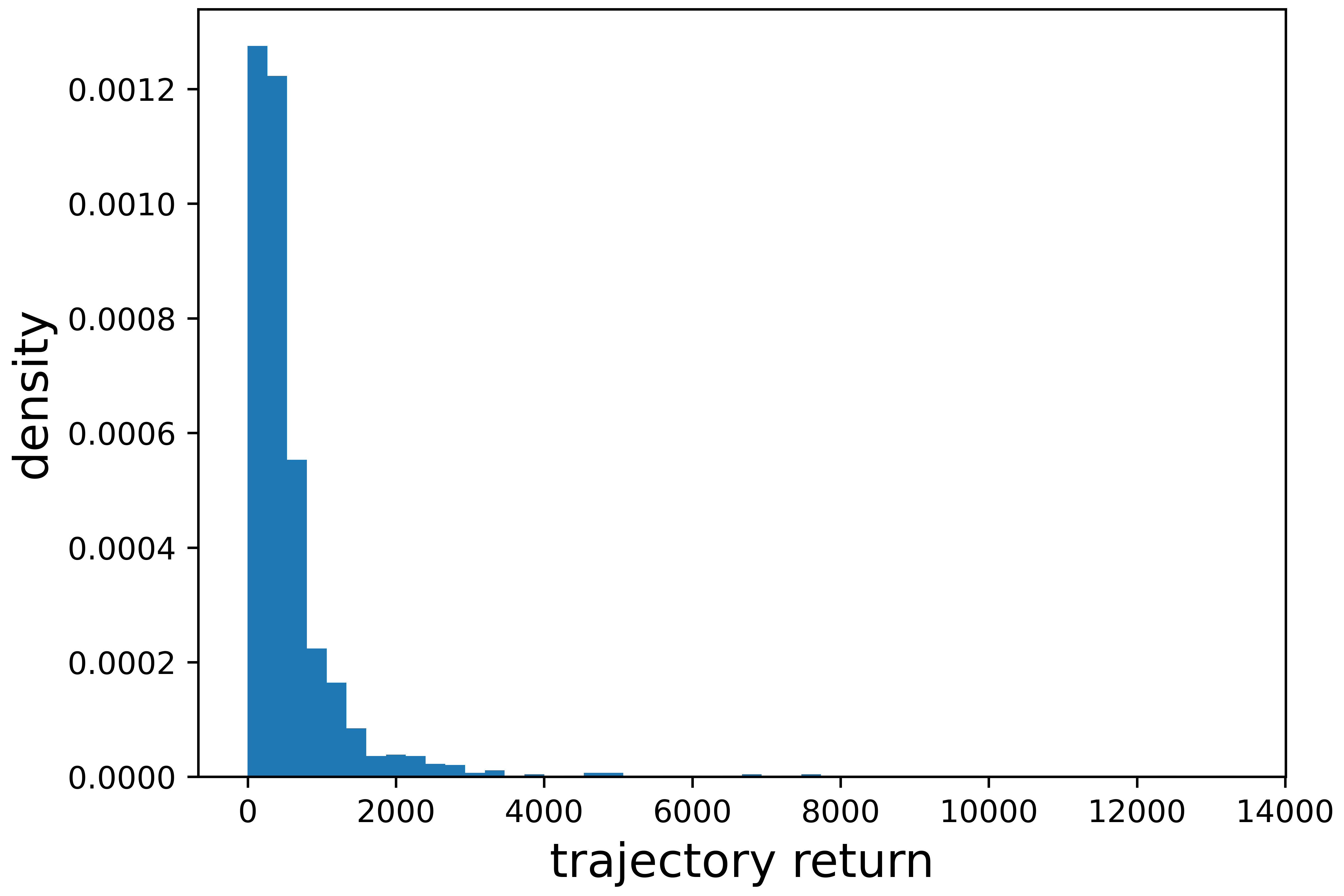}
         \caption{Hopper}
     \end{subfigure}
     \caption{Histogram of trajectory returns for the medium-replay versions of the three tasks. Unlike the medium-replay versions for hopper and walker2d, this version for halfcheetah is not dominated by poor trajectories.}
     \label{fig:medium_replay_return}
\end{figure*}

In \Cref{tab:performance}, on the medium-replay version of halfcheetah BRAC-v and CQL do not perform as poor as they are   on the medium-replay version of other two tasks. This works explain such observation with the distribution of trajectory returns in these datasets. As shown in \Cref{tab:datasets}, the normalized return of the medium-replay version for halfcheetah is much higher than the medium replay versions for walker2d and hopper. Moreover, as shown in \Cref{fig:medium_replay_return}, the medium-replay versions of walker2d and hopper are dominated by poor trajectories. The same version for halfcheetah, on the contrary, has much more good trajectories. As this version contains trajectories collected during training a RL agent, this indicates a faster convergence on halfcheetah than on other task. In consequence, this version for halfcheetah are generated by less policies than the same version for other tasks. 

\subsection{Results for Varying $K$}

\begin{table*}[t]
\centering
\caption{Normalized return of LBRAC-v for different $K$ on the D4RL datasets. The best value for $K$ is dataset-dependant.}
\scriptsize
\begin{tabular}{ccrrrrr}
\toprule
Task &Version & $K$=1           & $K$=5             & $K$=10            & $K$=15    & $K=20$             \\ \midrule
\multirow{4}{*}{halfcheetah} 
&random & 25.75$\pm$0.38 & 25.55$\pm$0.24 & 25.01$\pm$0.39 & \textbf{26.37$\pm$0.29} & 25.44$\pm$0.23 \\
&medium & 51.93$\pm$0.13 & \textbf{52.14$\pm$0.11} & 51.71$\pm$0.11 & 52.01$\pm$0.11 & 51.56$\pm$0.17 \\
&medium-replay & \textbf{48.41$\pm$0.14} & 48.16$\pm$0.18 & 47.94$\pm$0.12 & 47.99$\pm$0.17 & 47.89$\pm$0.09 \\
&medium-expert & 93.14$\pm$1.05 & 91.54$\pm$1.36 & 91.22$\pm$1.15 & \textbf{94.22$\pm$1.00} & 92.44$\pm$1.26 \\ \midrule

\multirow{4}{*}{walker2d} 
&random & \textbf{11.96$\pm$4.39} & 10.22$\pm$2.56 & 7.10$\pm$1.75 & 8.53$\pm$3.15 & 10.82$\pm$4.18 \\
&medium & 77.98$\pm$1.74 & 80.56$\pm$1.45 & 78.45$\pm$2.27 & 79.63$\pm$2.17 & \textbf{82.40$\pm$1.30} \\
&medium-replay & 61.93$\pm$5.08 & 65.53$\pm$3.45 & 63.79$\pm$4.21 & 74.20$\pm$3.74 & \textbf{76.18$\pm$4.47} \\
&medium-expert & 109.52$\pm$0.25 & 109.37$\pm$0.16 & 109.36$\pm$0.20 & 109.57$\pm$0.26 & \textbf{109.65$\pm$0.23} \\ \midrule

\multirow{4}{*}{hopper} 
&random & 9.43$\pm$0.49 & 8.96$\pm$0.32 & 8.75$\pm$0.66 & \textbf{9.63$\pm$0.27} & 8.46$\pm$0.16 \\
&medium & 97.73$\pm$2.56 & 96.46$\pm$1.52 & 100.20$\pm$1.03 & \textbf{100.50$\pm$0.57} & 96.16$\pm$2.52 \\
&medium-replay & \textbf{81.89$\pm$12.77} & 47.69$\pm$10.17 & 78.45$\pm$12.67 & 78.37$\pm$10.66 & 64.94$\pm$9.17 \\
&medium-expert & 82.06$\pm$14.33 & 73.27$\pm$13.50 & 82.07$\pm$11.69 & 76.92$\pm$7.84 & \textbf{98.08$\pm$6.39} \\ \bottomrule
\end{tabular}
\label{tab:varying_k_d4rl}
\end{table*}

\begin{table*}[t]
\centering
\caption{Relative return of LBRAC-v and for different $K$ on the heterogeneous-$k$ datasets.}
\scriptsize
\begin{tabular}{ccrrrrr}
\toprule
Task &\#Behavior Policy & $K$=1           & $K$=5             & $K$=10            & $K$=15  & $K$=20            \\ \midrule
\multirow{4}{*}{halfcheetah} 
&1 & 1.0129$\pm$0.0073 & 0.9988$\pm$0.0056 & \textbf{1.0244$\pm$0.0025} & 1.0056$\pm$0.0078 & 0.9922$\pm$0.0164 \\
&2 & \textbf{1.0106$\pm$0.0118} & 0.9686$\pm$0.0229 & 0.9914$\pm$0.0286 & 0.9741$\pm$0.0269 & 0.9707$\pm$0.0125 \\
&3 & 0.9160$\pm$0.0369 & 0.8757$\pm$0.0738 & 0.9608$\pm$0.0114 & 0.9578$\pm$0.0311 & \textbf{0.9709$\pm$0.0450} \\
&4 & 0.9035$\pm$0.0424 & 0.9574$\pm$0.0124 & 0.9643$\pm$0.0249 & 0.9616$\pm$0.0202 & \textbf{0.9761$\pm$0.0178} \\
&5 & 0.8754$\pm$0.0394 & 0.8632$\pm$0.0371 & \textbf{0.8856$\pm$0.0237} & 0.8666$\pm$0.0473 & 0.5129$\pm$0.0448 \\ \midrule

\multirow{4}{*}{walker2d} 
&1 & 1.0422$\pm$0.0094 & \textbf{1.0556$\pm$0.0044} & 1.0524$\pm$0.0026 & 1.0513$\pm$0.0032 & 1.0510$\pm$0.0016 \\
&2 & 1.0250$\pm$0.0187 & 0.9560$\pm$0.0566 & \textbf{1.0392$\pm$0.0183} & 1.0060$\pm$0.0079 & 1.0371$\pm$0.0205 \\
&3 & 0.7741$\pm$0.1871 & \textbf{0.8779$\pm$0.1516} & 0.7664$\pm$0.1224 & 0.6280$\pm$0.1160 & 0.5828$\pm$0.1943 \\
&4 & \textbf{0.9442$\pm$0.0355} & 0.8178$\pm$0.0680 & 0.8268$\pm$0.1004 & 0.7786$\pm$0.1298 & 0.7642$\pm$0.0965 \\
&5 & \textbf{1.0113$\pm$0.0389} & 0.8073$\pm$0.1039 & 0.8709$\pm$0.1251 & 0.9192$\pm$0.0784 & 0.8308$\pm$0.1683 \\ \midrule

\multirow{4}{*}{hopper} 
&1 & 0.9052$\pm$0.0315 & 0.9237$\pm$0.0217 & 0.9352$\pm$0.0338 & 0.9355$\pm$0.0186 & \textbf{0.9618$\pm$0.0147} \\
&2 & 0.9650$\pm$0.0364 & 0.9101$\pm$0.0504 & 0.9472$\pm$0.0590 & \textbf{0.9914$\pm$0.0197} & 0.9477$\pm$0.0445 \\
&3 & 1.0614$\pm$0.0099 & 1.0535$\pm$0.0142 & 1.0630$\pm$0.0080 & \textbf{1.0669$\pm$0.0050} & 1.0654$\pm$0.0101 \\
&4 & 1.0710$\pm$0.0451 & 1.0129$\pm$0.0447 & 1.1072$\pm$0.0133 & 0.9505$\pm$0.0949 & \textbf{1.1168$\pm$0.0128} \\
&5 & 1.0738$\pm$0.0508 & 1.0044$\pm$0.0354 & \textbf{1.1354$\pm$0.0146} & 0.9162$\pm$0.1295 & 0.9500$\pm$0.0947 \\ \bottomrule
\end{tabular}
\label{tab:varying_k_heterogeneous}
\end{table*}

Table~\ref{tab:varying_k_d4rl} shows the results of LBRAC-v for different $K$ on the D4RL datasets. The best value of $K$ is dataset dependent. For example, the best value for the medium version of halfcheetah is $K=5$, but for the random version it is $K=15$. 

\subsection{Additional Visualizations}
\begin{figure*}
     \centering
     \begin{subfigure}[b]{\textwidth}
         \centering
         \includegraphics[width=\textwidth]{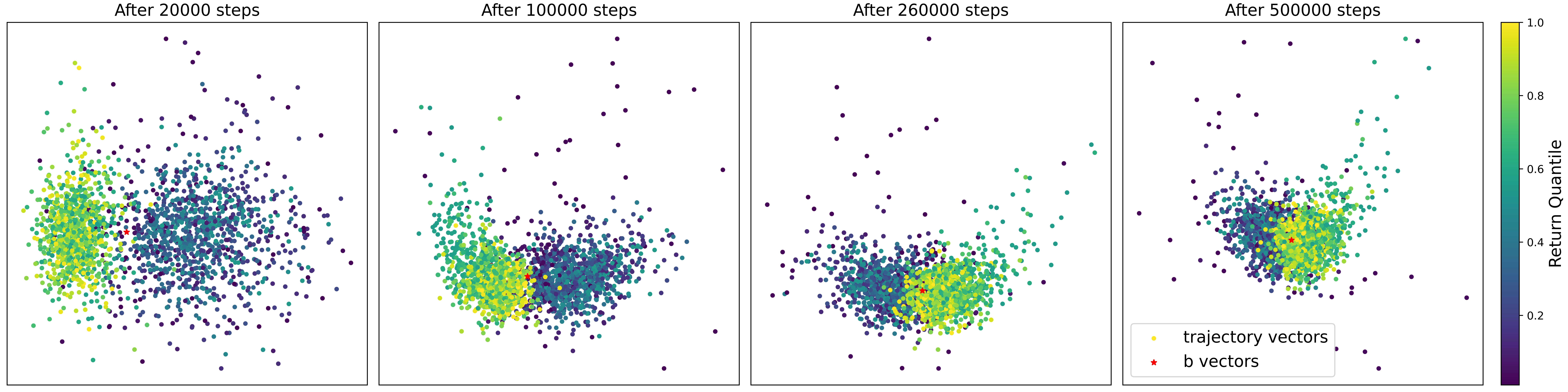}
         \caption{The proposed model for $K=1$.}
         \label{fig:vqvae_training_k=1}
     \end{subfigure}
     \begin{subfigure}[b]{\textwidth}
         \centering
         \includegraphics[width=\textwidth]{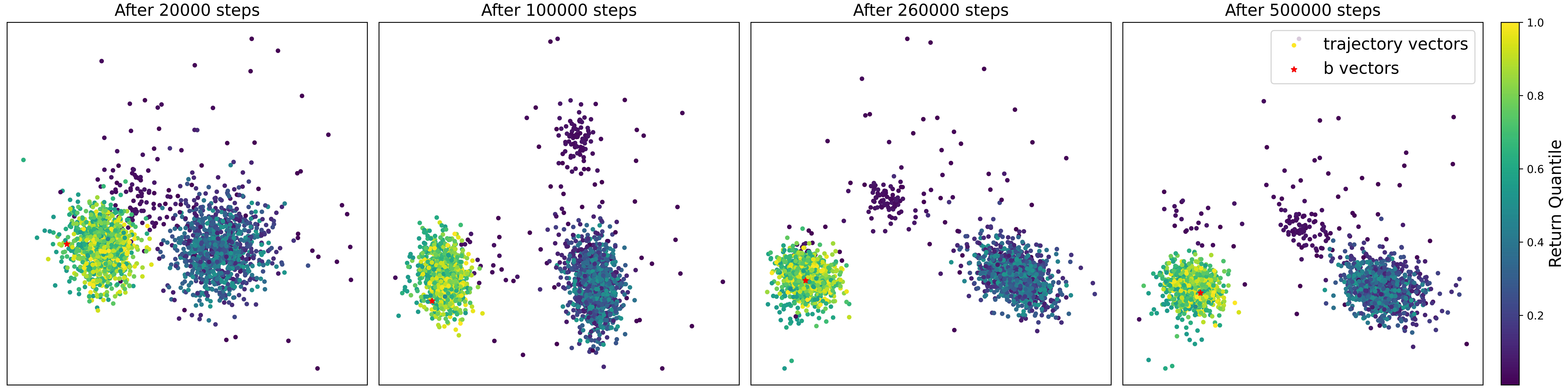}
         \caption{LBRAC-v for $K=1$.}
         \label{fig:lbrac_training_k=1}
     \end{subfigure}
     \begin{subfigure}[b]{\textwidth}
         \centering
         \includegraphics[width=\textwidth]{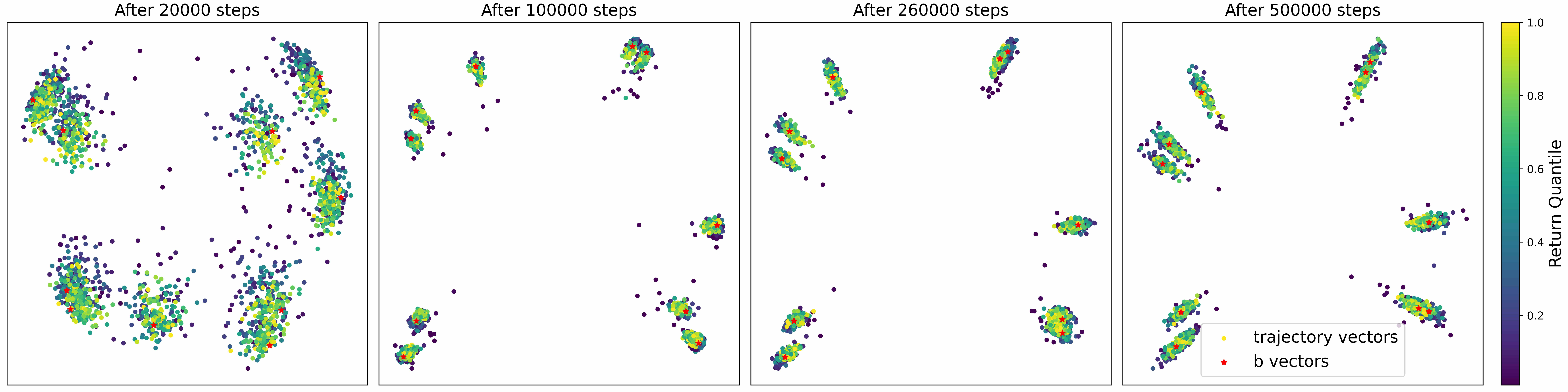}
         \caption{The proposed model for $K=10$.}
         \label{fig:vqvae_training_k=10}
     \end{subfigure}
     \begin{subfigure}[b]{\textwidth}
         \centering
         \includegraphics[width=\textwidth]{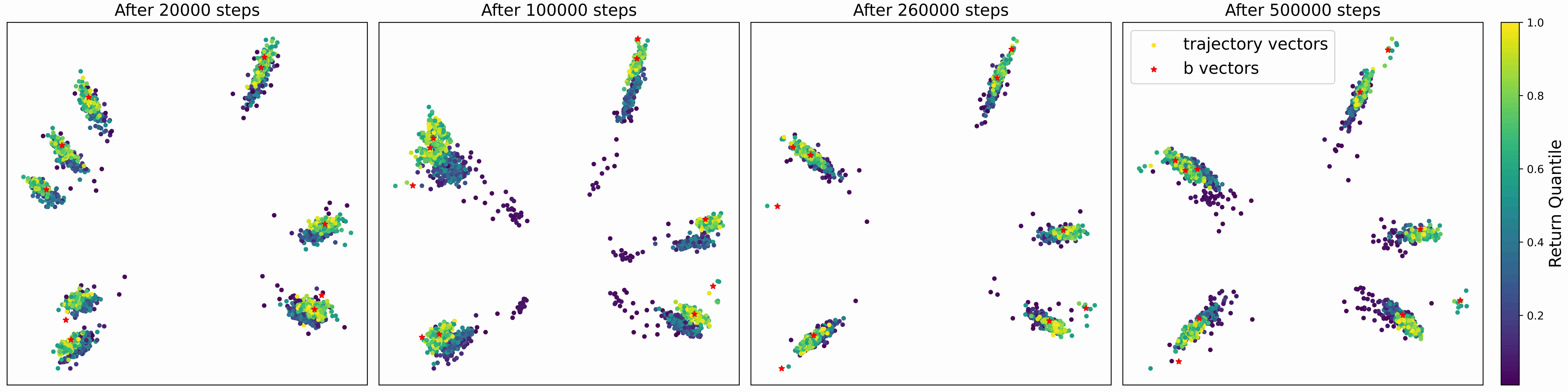}
         \caption{LBRAC-v for $K=10$.}
         \label{fig:lbrac_training_k=10}
     \end{subfigure}        
     \caption{Visualizations for the proposed model and LBRAC-v during training. }
     \label{fig:add_training}
\end{figure*}

\Cref{fig:vqvae_training_k=1}, \Cref{fig:lbrac_training_k=1}, \Cref{fig:vqvae_training_k=10} and \Cref{fig:lbrac_training_k=10} presents visualizations obtained during training the proposed model and LBRAC-v for $K=1$ and $K=10$.

\end{document}